\documentclass[10pt,twocolumn,letterpaper]{article}
\usepackage{iccv}
\usepackage{times}
\usepackage{epsfig}
\usepackage{graphicx}
\usepackage{amsmath}
\usepackage{amssymb}
\usepackage{multirow}
\usepackage{ctable}
\usepackage{array}
\usepackage{graphicx,subfigure}
\usepackage{amsmath}
\usepackage{amssymb}
\usepackage{enumitem}
 
\usepackage{url}
\usepackage{subfigure}
\usepackage[linesnumbered,ruled]{algorithm2e}

\graphicspath{{fig/}}

\usepackage{color}
\usepackage{ulem}
\definecolor{crimson}{rgb}{0.86, 0.08, 0.24}
\ifx \submission \undefined

\else

\fi

\usepackage[pagebackref=true,breaklinks=true,letterpaper=true,colorlinks,bookmarks=false]{hyperref}

\iccvfinalcopy 

\def\iccvPaperID{9999} 

\ificcvfinal\fi

\begin{document}

\title{Self Adversarial Training for Human Pose Estimation}

\author{Chia-Jung Chou\,, \, Jui-Ting Chien\,, and \, Hwann-Tzong Chen\\
Department of Computer Science\\
National Tsing Hua University, Taiwan\\
{\tt\small \{jessie33321, ydnaandy123\}@gmail.com}
\qquad
{\tt\small htchen@cs.nthu.edu.tw}
}

\maketitle


\begin{abstract}
This paper presents a deep learning based approach to the problem of human pose estimation. We employ generative adversarial networks as our learning paradigm in which we set up two stacked hourglass networks with the same architecture, one as the generator and the other as the discriminator. The generator is used as a human pose estimator after the training is done. The discriminator distinguishes ground-truth heatmaps from generated ones, and back-propagates the adversarial loss to the generator. This process enables the generator to learn plausible human body configurations and is shown to be useful for improving the prediction accuracy. 
\end{abstract}

\section{Introduction}

Human pose estimation from a single image is a challenging problem due to the limited information of 2D images and the large variations in configuration and appearance of body parts. Early work often tackles the problem using graphical models \cite{AndrilukaRS09,FelzenszwalbMR08,JohnsonE11} and random field inference \cite{LadickyTZ13,RamakrishnaMHBS14} with handcrafted image features. Despite the improvements made by those intriguing designs of models and algorithms, the bottleneck seems to be the lack of effective feature representations that are capable of characterizing different levels of visual cues and accounting for the varieties in appearance of people. 

The situation has been changed along with the popularity of deep learning in computer vision. Deep neural nets have the ability to learn better feature representations. For example, a recent approach, {\em stacked hourglass network}~\cite{NewellYD16}, achieves state-of-the-art performance without the use of hand designed priors or graphical-model-style inference. The well-designed architecture, which supports repeated bottom-up, top-down inference across scales for large receptive field, helps the model to capture some correlations among human body parts. However, the model might predict human pose with implausible configuration due to severe occlusion or overlapping with other people nearby. In these situations, the model is forced to find some similar features which might be in the background or belong to another person. 
These challenging cases are much easier for human vision to recognize. Humans have the concepts of the structure and constraint of body parts, and are also good at associating these concepts with observed image features. Inspired by the success of {\em generative adversarial networks} on many topics, we incorporate a discriminator to take charge of checking the structural constraints of human body. We maintain the original pose estimator as the generator to capture important image features. It is worth noting that the architectures of our discriminator and generator are exactly the same. We use the adversarial training strategy to enable the discriminator to distinguish implausible poses and simultaneously to guide the generator. After the training is done, the generator is used as a pose estimator and the discriminator can be removed.

\begin{figure}[t]
\centering
    \subfigure[] { \includegraphics[width=0.185\textwidth]{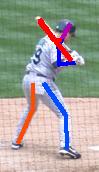} }
    \subfigure[] { \includegraphics[width=0.185\textwidth]{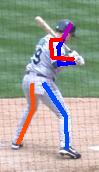} }
    \caption{\label{fig:motiv} Motivation. (a) A deep network may produce incorrect estimations due to occlusion. (b) After incorporating adversarial training, the structural constraints of human body parts can be learned.}
 \end{figure}

The main contribution of this work is two folds: First, we design a deep ConvNet model to learn the structure and configuration of human body parts via adversarial training. The training techniques of generative adversarial networks are used to train the proposed model for solving the human pose estimation problem. Second, we evaluate our method using LSP, MPII, and LIP datasets, with detailed analysis on the effects of different components in our design, and the experimental results show improved accuracy on all of those datasets.

\section{Related Work}

\subsection{Human Pose Estimation}

Many recent methods on human pose estimations use deep neural nets to predict the keypoints of human body in an image. DeepPose \cite{ToshevS14}, one of the earliest deep-learning based approaches to human pose estimation, formulates the pose estimation problem as a regression problem using a standard convolutional architecture, and its performance is higher than classical approaches \cite{AndrilukaRS09,DantoneGLG13,FelzenszwalbMR08,JohnsonE10,PishchulinAGS13,YangR11}. Latest methods mostly aim to predict structural outputs, usually called heatmaps or support maps that characterize the probabilities of observing each keypoint at different locations. The exact location of a keypoint is further estimated by finding the maximum in an aggregation of heatmaps. Compared with direct-regression methods, heatmap-based methods better leverage the distributed properties of convolutional networks and are considered more suitable for training.

Some works incorporate graphical models, \eg CRF, MRF, which may be used as a post-processing step \cite{ChenY14} or embedded into the network for end-to-end training \cite{ChuYOMYW17,YangOHW16}.
Powerful CNN architectures have been developed to capture the important cues and evidences of human parts. In \cite{WeiRKS16} and \cite{NewellYD16}, a multi-stage scheme is employed to make the receptive field large enough for learning the long-range spatial relationships. Also, intermediate supervision is used to produce intermediate confidence maps and let them be refined through different stages. Several recent methods focus on solving the multi-person pose estimation problem. The methods of \cite{InsafutdinovPAA16,PishchulinITAAG16} estimate poses of multiple people in a single image. They use deep networks to generate keypoint candidates and run integer linear programming (ILP) to group joints candidates for each person. The approach of Cao \etal \cite{CaoSWS16} predicts the multi-person keypoint heatmaps and the part affinity fields, and then uses a greedy algorithm to group the joints that belong to the same person.

\subsection{Generative Adversarial Networks}

{\em Generative adversarial networks} (GANs) flourish in generating natural images such as human faces and indoor scenes. With the introduction by Goodfellow \etal \cite{GoodfellowPMXWOCB14}, the two-player minimax game allows unsupervised training of generative models and avoids the blur effect of using variational autoencoders. However, people concern about GANs being unstable and hard to train. Radford \etal \cite{RadfordMC15} introduce DCGAN, an all convolutional architecture which is easier to train. They propose some elements to increase the model stability such as eliminating the fully connected layer and employing batch normalization to prevent from losing diversity, \ie, mode collapsing.
DCGAN uses an effective network configuration to make the training of GAN more feasible. 

Recently, Arjovsky \etal \cite{ArjovskyCB17} propose Wasserstein GAN (WGAN), which does not require a special network design like DCGAN. WGAN uses the Wasserstein distance to replace the original loss function in GAN and solves the unreliable gradient problem in the original GAN. Using Wasserstein distance also provides an estimate of the quality of the generated samples. However, since WGAN satisfies the K-Lipschitz constraint by weight clipping, it pushes weights towards two values (the extremes of the clipping range) and is hard to tune the clipping parameters. 
Gulrajani \etal \cite{GulrajaniAADC17} replace the weight clipping strategy by gradient penalty. Gradient penalty is an additional term in the loss function that directly enforces the discriminator's gradient norm around $K$. The result shows that the improved training strategy of \cite{GulrajaniAADC17} is much faster and more stable than WGAN. Another branch of work uses autoencoders as discriminators such as EBGAN \cite{ZhaoML16}. EBGAN aims to match the autoencoder loss distribution while typical GANs try to match the data distribution. EBGAN still suffers from the same problem of classical GANs. Inherited from \cite{ZhaoML16}, Berthelot \etal \cite{BerthelotSM17} present an equilibrium term, which is based on {\em proportional control theory}, to balance the discriminator and the generator. It also provides a convergence measure that can be used to determine if the model has collapsed or reached its final state.

Due to the success of GAN on generating images, it also draws attention to the field of supervised learning. The concept of {\em conditional GAN} \cite{MirzaO14} is introduced for incorporating class information. Several methods combine the conditional GAN loss and the L1 or L2 distance between generated data and ground-truth data. The methods of \cite{IsolaZZE16,LedigTHCATTWS16,PathakKDDE16} use this solution to perform tasks of super-resolution, image inpainting, and image translation. They get promising results with respect to human vision. The examples described above are still all about generating natural images. They either generate a whole image based on certain constraints or generate an image patch. Another type of task is about generating heatmaps of labels as in semantic segmentation \cite{LucCCV16}, saliency \cite{PanCMOTSN17} or human pose estimation \cite{ChenSWLY17}. Adding the adversarial training strategy to this type of task seems to bring some benefits to it. In our work, we also try to use adversarial training techniques \cite{BerthelotSM17}.
to improve the performance of pose estimator.

\section{Adversarial Training with the Stacked Hourglass Networks}

\begin{figure*}[tb]
    \centering
    \includegraphics[width=1\textwidth]{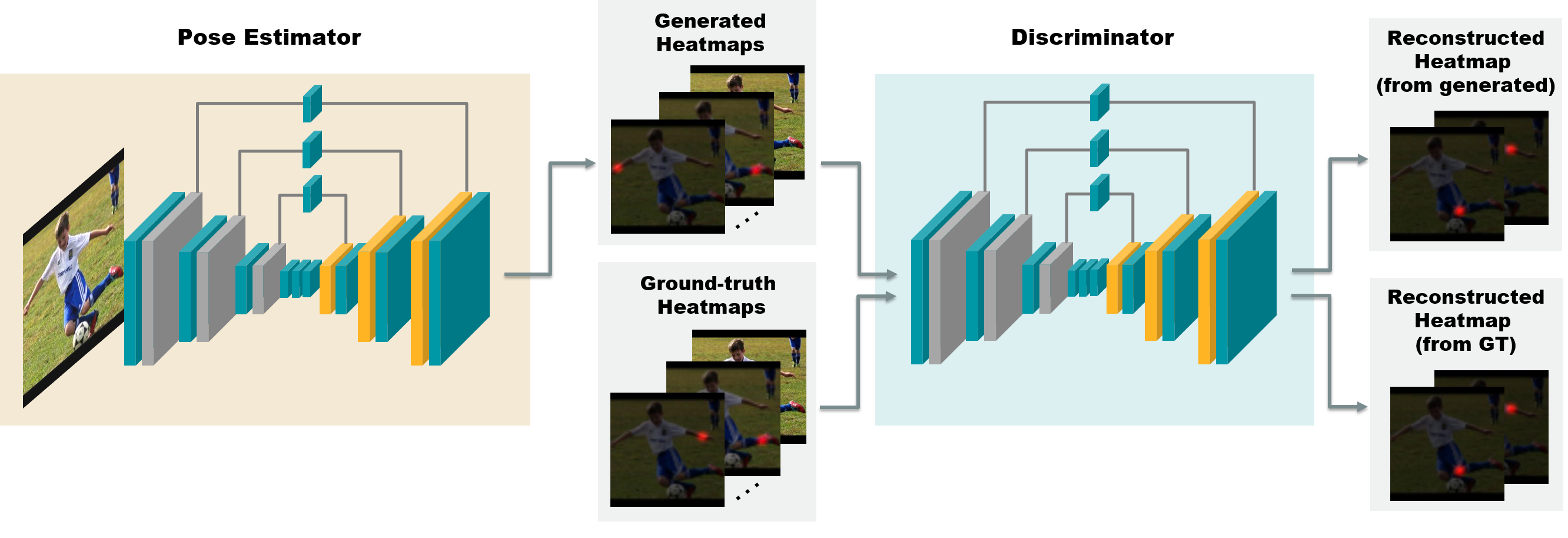}
    \caption{ The framework of our adversarial networks. We incorporate a ConvNet-based pose estimator as the generator (on the left) with a discriminator (on the right) that aims to distinguish the generated heatmaps from the ground-truth heatmaps by reconstructing the input heatmaps. The generator and the discriminator have the same architecture.}
    \label{fig:framework}
\end{figure*}

Our model splits into two networks, the generator and the discriminator. The first network, {\em generator}, is a fully convolutional network with residual blocks and a conv-deconv architecture. After feeding forward through the generator, we get a set of heatmaps that indicate the confidence score at every location for each keypoint. The second network, {\em discriminator}, has the same architecture as the generator but it encodes the heatmaps along with the RGB image and decodes them into a new set of heatmaps in order to distinguish real heatmaps from fake ones. The framework of our model is illustrated in Fig.~\ref{fig:framework}.

\subsection{Generator}
The goal of the generative network is to learn a mapping from a color image to keypoint heatmaps. The deep convolutional architecture allows itself to learn contextual feature representation from the input images. Additionally, the adversarial loss from the discriminative network is introduced and combined with the error between the generated heatmaps and the ground-truth heatmaps. This process enables the generator to learn not only the features and spatial dependencies from images but also the plausible human body configurations.

\subsubsection{Network Architecture}
\begin{figure*}[t]
	\centering
	\includegraphics[width=0.88\textwidth]{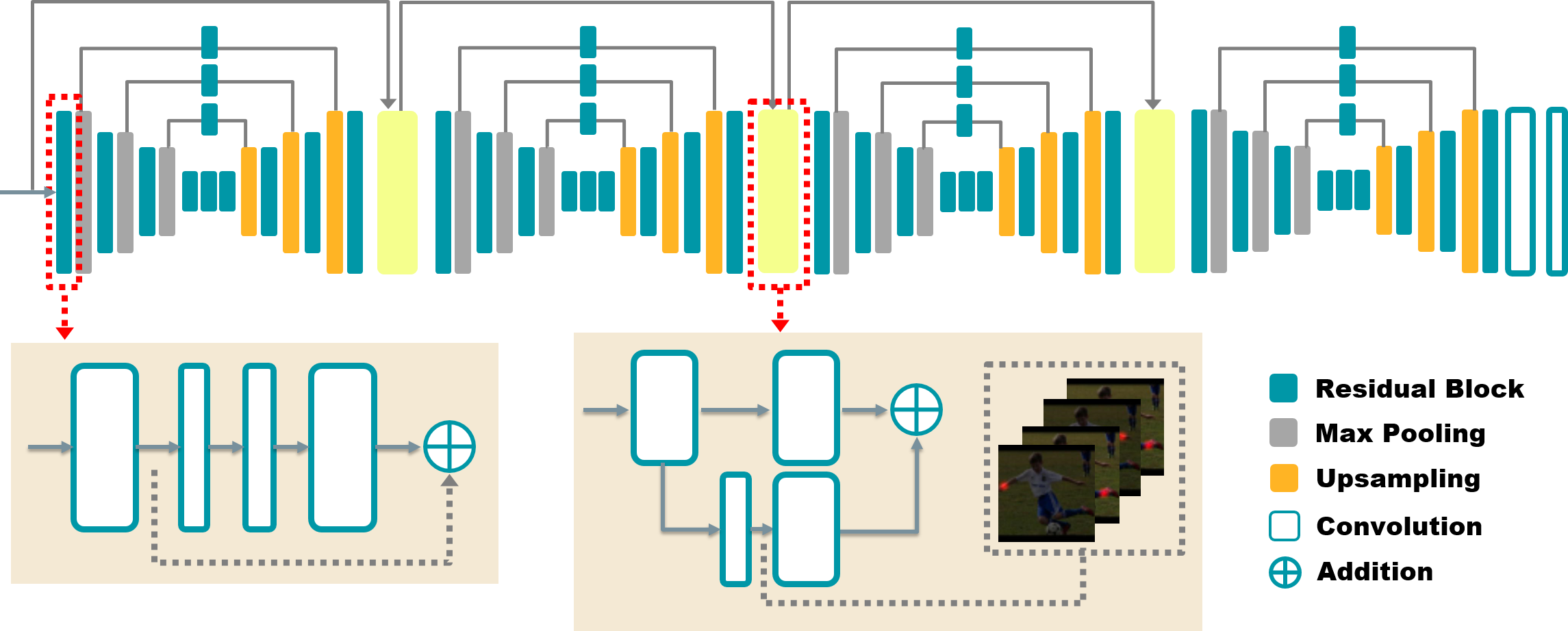}
    \caption{The architecture of 4-stack hourglass. The hourglass module consists of residual blocks (zoomed-in at bottom-left), pooling layers, upsampling layers, and skip connections. Between each pair of consecutive hourglass stacks, there is a transition block (yellow box) which produces intermediate heatmaps and adds them to the main trunk of the network.}
    \label{fig:generator}
\end{figure*}

We use the state-of-the-art hourglass architecture \cite{NewellYD16} as our base network. It is a fully convolutional network with residual modules as its building blocks. The network starts with an initial process of a $7\times7$ convolution with stride 2, followed by several residual modules and max-pooling layers. The initial process reduces the resolution of the feature maps from 256 to 64. Then, a sequence of hourglass modules are stacked to predict the keypoint heatmaps. A single hourglass module is a bottom-up and top-down design to extract the features at every scale. For human pose estimation, it is essential to explore both the local evidence, such as a small region around the wrist, and the long-range relationships between joints. To maintain the information and to integrate global and local context concurrently, skip connections are used, and features at each resolution can be better preserved. A 4-stack hourglass architecture is shown in Fig.~\ref{fig:generator}

\subsubsection{Training the Generator} \label{subsec:first_training}

Training the generator is done by back-propagating the loss $\mathcal{L}_{\mathrm{MSE}}$ from generator itself and the adversarial loss $\mathcal{L}_{\mathrm{adv}}$ from the discriminator. 

\begin{figure*}[t]
	\centering
	\includegraphics[width=0.9\textwidth]{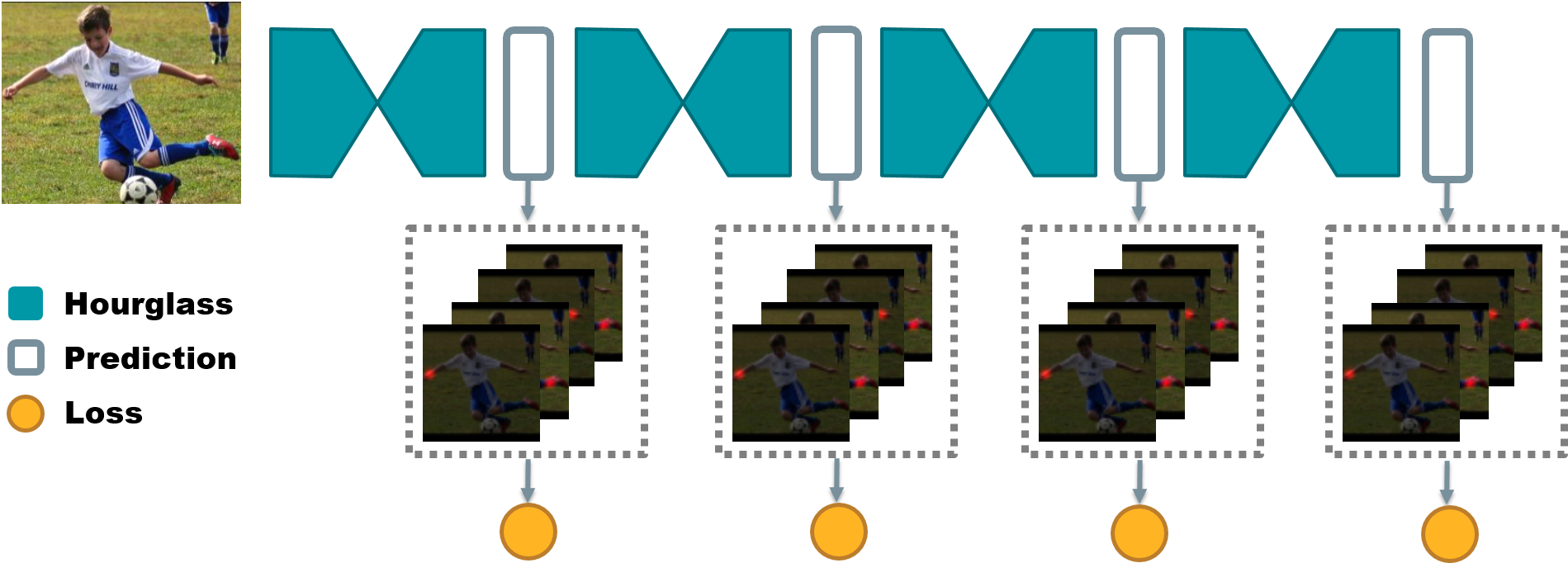}
    \caption{An illustration of intermediate supervision. The mean squared error (MSE) loss is applied at the end of each hourglass module.}
    \label{fig:mse}
\end{figure*}

The generator consists of $N$ stacks of hourglass modules. The expected output of each hourglass module contains $M$ heatmaps, each of which is a $64\times64$ map with a Gaussian peaked at the ground-truth location of the $j$th joint. The supervision is conducted at the end of each hourglass. The loss from the generator itself can be expressed as
\begin{equation}
	\begin{aligned}
		\mathcal{L}_{\mathrm{MSE}}\;=\; \sum\limits_{i=1}^{N}\sum\limits_{j=1}^M (C_{ij} -\hat{C}_{ij})^2 \,,
	\end{aligned}
    \label{G1}
\end{equation}
where $C_{ij}$ is the ground-truth heatmap of $j$th joints at the $i$th stack, and $\hat{C}_{ij}$ is the generated heatmap. We calculate the mean square error between them to enforce the generator to learn the image features that are important for localizing the keypoints. In early stacks, local evidence is used since the receptive field is restricted to a small area. In later stacks, long-range spatial relationships will be considered since the receptive field has been enlarged through the numerous sequential convolutional operations. This training scheme is illustrated in Fig.~\ref{fig:mse}.

In addition to the traditional supervised loss described above, we add an adversarial loss, which can urge the generator to produce reasonable poses. The adversarial loss from the discriminator can be expressed as
\begin{equation}
	\begin{aligned}
		\mathcal{L}_{\mathrm{adv}}\;=\; \sum\limits_{j=1}^M (\hat{C}_{j}-D(\hat{C}_{j}, X))^2 \,,
	\end{aligned}
    \label{G2}
\end{equation}
where $\hat{C}_{j}$ is the output heatmaps of the generator's last hourglass stack, $D$ is the discriminator, and $X$ is an input image. The loss computes the error between generated heatmaps and reconstructed heatmaps. The detail of this equation will be explained in the next section.

The total loss for generator is defined by 
\begin{equation}
	\begin{aligned}	
		\mathcal{L}_{G}\;=\; \mathcal{L}_{\mathrm{MSE}} \; + \;&\lambda_{G}\,\mathcal{L}_{\mathrm{adv}}\,,
	\end{aligned}
\end{equation}
where $\lambda_{G}$ is a hyperparameter to control the weight of adversarial loss.

\subsection{Discriminator}

The objective of the discriminator is to distinguish real data from generated data. The input of the discriminator contains either ground-truth heatmaps or generated heatmaps, and both of them are concatenated with the corresponding color image of the person. From the input pair, the discriminator should learn whether the pose described by the heatmaps is correct and corresponds to the person in the input color image. The discriminator attempts to reconstruct a new set of heatmaps. The qualities of the reconstructed heatmaps are determined by how they are similar to the input heatmaps, following the same notion as autoencoder. The loss is computed as the error between the input heatmaps and the reconstructed heatmaps.

\subsubsection{Training the Discriminator}
For each training image, the generated and ground-truth heatmaps will be fed to the discriminator separately. Two sets of heatmaps will be reconstructed for computing $\mathcal{L}_{\mathrm{real}}$ and $\mathcal{L}_{\mathrm{fake}}$. In other words, at each iteration, the discriminator is updated using the accumulated gradient, which is computed with respect to $\mathcal{L}_{\mathrm{real}}$ and $\mathcal{L}_{\mathrm{fake}}$.

When the input comprises ground-truth heatmaps, the discriminator is trained to recognize it and reconstruct a similar one, \ie, to minimize the error between the ground-truth heatmaps and the reconstructed ones. On the other hand, if the input comprises generated heatmaps, the discriminator is trained to reconstruct totally different ones, \ie, to drive the error between the generated heatmaps and the reconstructed ones as large as possible. The loss is expressed as 
\begin{equation}
	\begin{aligned}	
		\mathcal{L}_{\mathrm{real}}\;=&\; \sum\limits_{j=1}^M (C_{j}-D(C_{j}, X))^2, \\
        \mathcal{L}_{\mathrm{fake}}\;=&\;\sum\limits_{j=1}^M (\hat{C}_{j}-D(\hat{C}_{j}, X))^2,\\
        \mathcal{L}_{D}\;=&\;\mathcal{L}_{\mathrm{real}}-k_{t}\,\mathcal{L}_{\mathrm{fake}}\,.
	\end{aligned}
    \label{D_loss}
\end{equation}
The discriminator is optimized by the per-pixel loss $\mathcal{L}_{D}$. Given a set of heatmaps, which can refer to a particular pose, the discriminator will give a value to each pixel. The value is the error between the input and output heatmaps of the discriminator. The value means how good the confidence of this pixel is, in the discriminator's opinion. 
For example, if the confidence of the right knee is high nearby the left knee, a well-trained discriminator will produce a heatmap of the right knee that has a larger error at the location of left knee. 

Since the discriminator is like a critic, it offers detailed `comments' on the input heatmaps and suggests which parts in the heatmaps do not yield a real pose. This is different from the conventional GAN, which only judges the whole input being good or bad.  

As mentioned in many papers, GAN is unstable and hard to train since the network easily collapses when the discriminator gets too good too quickly. Inspired by \cite{BerthelotSM17}, we use a variable $k_{t}$ to control the balance between generator and discriminator. The variable is updated at every iteration $t$. The adaptive term $k_{t}$ is defined by 
\begin{equation}
	\begin{aligned}	
		k_{t+1}\;=\;k_{t}+\lambda_{k}\,(\gamma\,\mathcal{L}_{\mathrm{real}}-\mathcal{L}_{\mathrm{fake}}) \,,
	\end{aligned}
\end{equation}
where $k_{t}$ is bounded between $0$ and $1$, and $\lambda_{k}$ is a hyperparameter. As in Eq.~(\ref{D_loss}), $k_{t}$ means how much emphasis is put on $\mathcal{L}_{\mathrm{fake}}$. When the generator gets better than the discriminator,  \ie, $\mathcal{L}_{\mathrm{fake}}$ is smaller than $\gamma\,\mathcal{L}_{\mathrm{real}}$, the generated heatmaps are real enough to fool the discriminator. Hence, $k_{t}$ will increase, to make the term  $\mathcal{L}_{\mathrm{fake}}$ more dominant, and thus the discriminator will be trained more on recognizing the generated heatmaps. The proportion it accelerates to train on $\mathcal{L}_{\mathrm{fake}}$ is according to how far the discriminator falls behind the generator, \ie, $\gamma\,\mathcal{L}_{\mathrm{real}}-\mathcal{L}_{\mathrm{fake}}$. Similarly, when the discriminator gets better than the generator, $k_{t}$ decreases, to slow down the training on $\mathcal{L}_{\mathrm{fake}}$ so that the generator can keep up with it.

\begin{figure*}[tb]
    \centering
    \includegraphics[width=0.925\textwidth,height=0.45\textwidth]{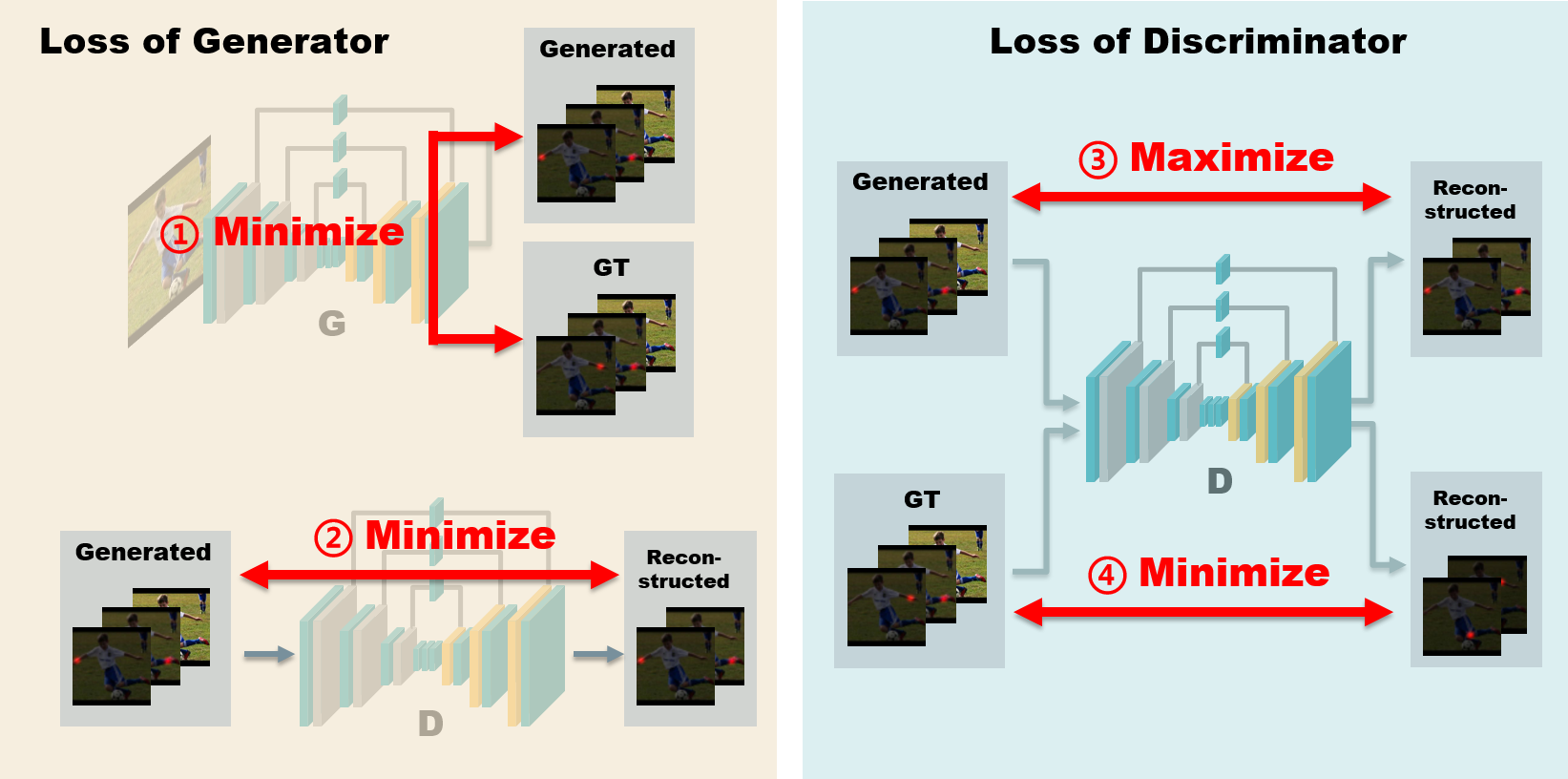}
    \caption{ Summary of $\mathcal{L}_{G}$ and $\mathcal{L}_{D}$. Losses in the orange box are used to update the generator. Losses in the blue box are used to update the discriminator.}
    \label{fig:loss_summary}
\end{figure*}


\subsection{Adversarial Training}

Based on {\em generative adversarial networks} (GANs), our training scheme is supervised learning plus a two-player game. The terms $\mathcal{L}_{\mathrm{fake}}$ in Eq.~(\ref{D_loss}) and Eq.~(\ref{G2}) have the same value except for the sign. The generator aims to minimize the distance between $\hat{C}$ and $D(\hat{C}, X)$ while the discriminator tries to maximize it. This is the adversarial part of this learning procedure. To distinguish poses, the discriminator seeks to capture the essential factor of real pose distribution during the process of reconstruction. At the same time, the generator seeks to produce higher-quality heatmaps of human pose so that it can deceive the discrminator and pass the inspection to let discriminator reconstruct similar heatmaps. In addition to the unsupervised training game, we preserve the traditional supervised learning to make the generator learn quicker and prevent the network from collapsing. Algorithm~\ref{algo} summarizes the adversarial training process.

\begin{algorithm}
    \SetKwInOut{Input}{Input}
    \SetKwInOut{Output}{Output}
    \SetKwRepeat{Do}{do}{while}
    \Input{An image $X$ of a person and the corresponding ground-truth heatmaps $C$}
	\Do{$\hat{C}$ still improves}
    {
      Forward discriminator by $D(C, X)$ 
      \\ Compute gradient $\nabla f_{D}$ \wrt Eq.~(\ref{D_loss})
      \\ Forward generator by $\hat{C} = G(X)$ 
      \\ Compute gradient $\nabla f_{G}$ \wrt Eq.~(\ref{G1})
      \\ Forward discriminator by $D(\hat{C}, X)$
      \\ Accumulate gradient $\nabla f_{D}$ \wrt Eq.~(\ref{D_loss})
	  \\ Update discriminator with $\nabla f_{D}$
      \\ Accumulate gradient $\nabla f_{G}$ \wrt Eq.~(\ref{G2})
	  \\ Update generator with $\nabla f_{G}$
    }
    \caption{The adversarial training process.}
    \label{algo}
\end{algorithm}

\subsubsection{Inference}

After the training is done, the discriminator can be removed. We use the generated heatmaps $\hat{C} = G(X)$ to infer the final result. To stabilize the predictions, we evaluate both the original image and its flipped version, and average their output heatmaps. As in the training phase, the output heatmap size of a joint is $64\times64$. We first extract the location with the largest confidence score in each joint's heatmap. Then, we transform the location back to the original coordinate space with respect to the input image size.

\section{Experiments}

\subsection{Datasets}

We evaluate our method on three benchmark datasets, Leeds Sports Pose Dataset (LSP) \cite{JohnsonE10}, MPII Human Pose Dataset \cite{AndrilukaPGS14}, and Look Into Person Dataset (LIP) \cite{GongLSL17}. In the following experiments, we use the same preprocessing and data augmentation settings. We randomly flip an input image horizontally, rotate it by an angle in $[-30,30]$ degrees, and scale it with factors in $[0.75,1.25]$. During testing, we scale the LSP and LIP images uniformly across the whole datasets to make the person a suitable size in the image. For MPII images, we use the scale and center annotations provided with the images. We implement our methods using Torch7 libraries for deep learning. We set a batch size of $6$ and train the network from scratch using the RMSprop optimization algorithm. The experiments are performed on a Titan X GPU. 
\begin{itemize}

\item \textbf{Leeds Sports Pose Dataset} (LSP):

Our results of LSP are trained on the LSP plus LSP-extended dataset. LSP consists of $11{,}000$ poses for training and $1{,}000$ for testing. The images are gathered from Flickr and contain people who are doing sports such as baseball, parkour, tennis, and so on. Each image is annotated with $14$ keypoint locations. To make it easier to integrate with other datasets, we calculate the center and scale of annotated person and use it at the training phase. The label of this dataset is a little noisy since some occluded joints may not have location information or the location might be wrong. The noisy labels and the variations in poses of humans doing sports make this dataset quite challenging.

\item \textbf{MPII Human Pose Dataset} (MPII) :

MPII dataset contains about $25{,}000$ images and over $40{,}000$ annotated people. These data are divided into $30{,}000$ images for training and $10{,}000$ images for testing. Each person is annotated with $16$ joints. The images are extracted from YouTube videos where the contents are everyday human activities. In comparison with other pose datasets, MPII has richer information such as activity label and fully unannotated video frames, and has higher image resolution. We only use keypoint locations during training.
　
\item \textbf{Look Into Person Human Pose Dataset} (LIP) :

LIP is the newest and largest dataset for human pose estimation. It contains $50{,}000$ images with $19$ semantic human part labels and $16$ human keypoints. In the following experiments, we only use keypoints information. The dataset divides into $30{,}462$ images for training set, $10{,}000$ images for validation set, and $10{,}000$ for test set. The images may contain full body, half body, or part of body, with heavy occlusions and of low resolution. The dataset is also used in CVPR 2017 workshop `Visual Understanding of Humans in Crowd Scene' and the first `Look Into Person (LIP) Challenge'.

\end{itemize}

\subsection{Evaluation Metrics}

The evaluations are based on two metrics. We use PCK to measure performance on LSP and LIP. For MPII, we use PCKh.

\begin{itemize}
\item \textbf{Percentage of Correct Keypoints} (PCK) \cite{YangR11}:

PCK reports the percentage of correct detection that falls within a tolerance range. The tolerance range is a fraction of torso size. The equation can be expressed as
\begin{equation}
	\frac{\| \mathbf{y}_i - \hat{\mathbf{y}}_i \| _2}{\| \mathbf{y}_{\mathrm{lhip}} - \mathbf{y}_{\mathrm{rsho}} \| _2} \leq r \,,
\end{equation}
where $\mathbf{y}_i$ is the ground-truth location of the $i$th keypoint and $\hat{\mathbf{y}}_i$ is the predicted location of the $i$th keypoint. The fraction $r$ is bounded between $0$ and $1$.

\item \textbf{Percentage of Correct Keypoints with respect to head} (PCKh) \cite{AndrilukaPGS14}:

PCKh is almost the same as PCK except for the tolerance range is a fraction of head size.
\end{itemize}

\subsection{Results}

\begin{figure*}
    \centering
    \includegraphics[width=0.95\textwidth]{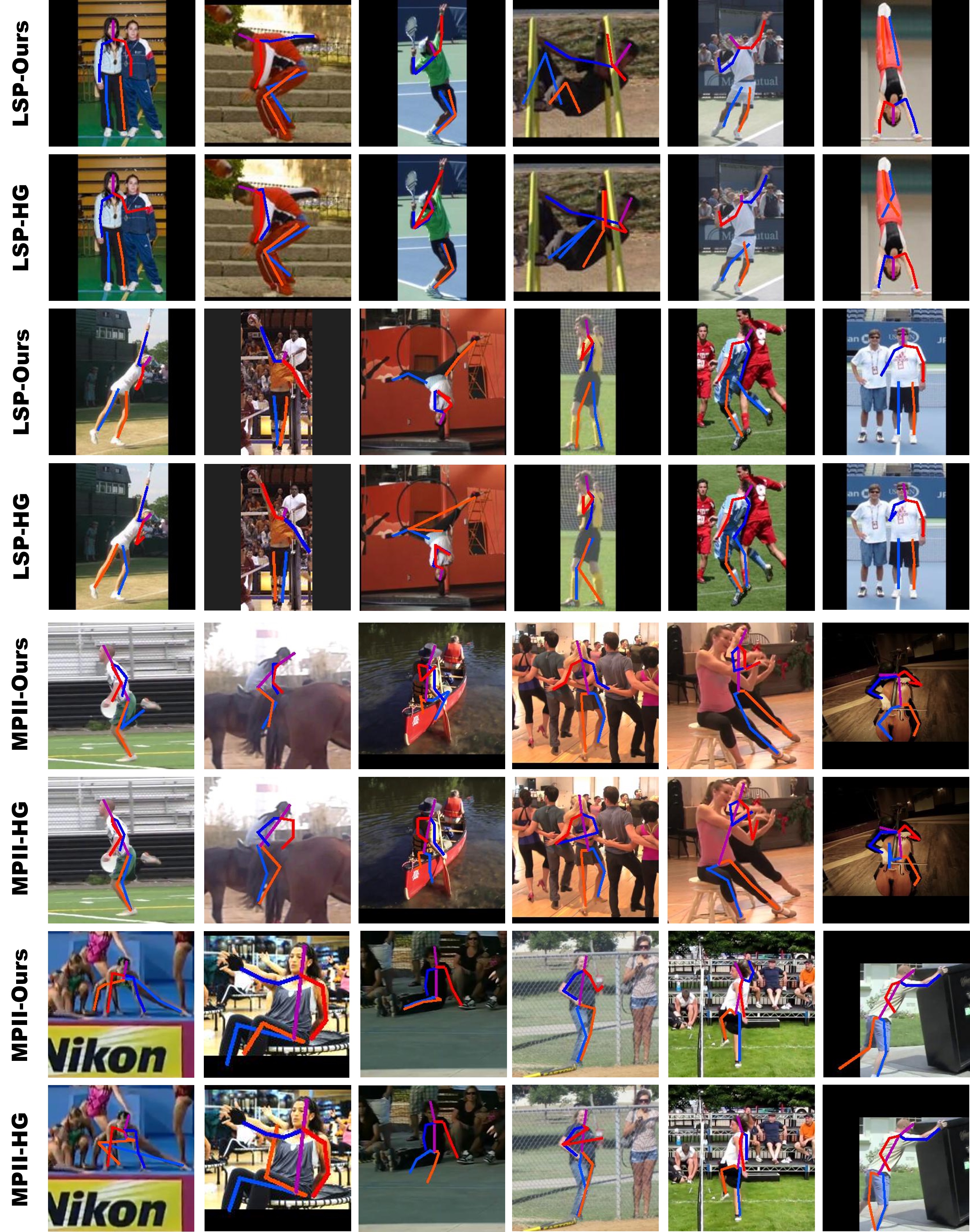}
    \caption{ Qualitative results. The red and orange lines indicate the left side, and the blue line indicates the right side. Our method can generate more plausible and structural poses than \cite{NewellYD16}.}
    \label{fig:qualitative}
\end{figure*}

\begin{figure*}
\centering
    \subfigure[] { \label{fig:improved_heatmaps_a}
    \includegraphics[width=0.85\textwidth]{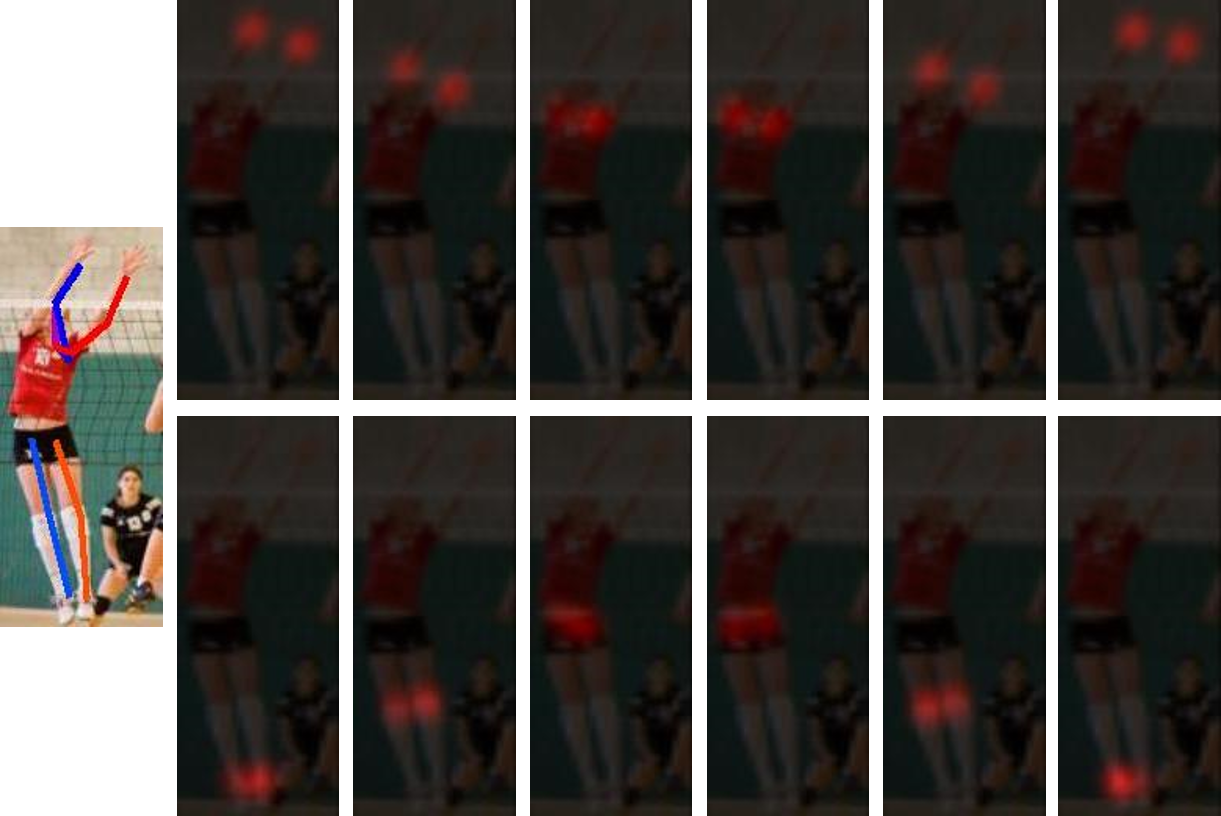} }
    \subfigure[] { \label{fig:improved_heatmaps_b}
    \includegraphics[width=0.85\textwidth]{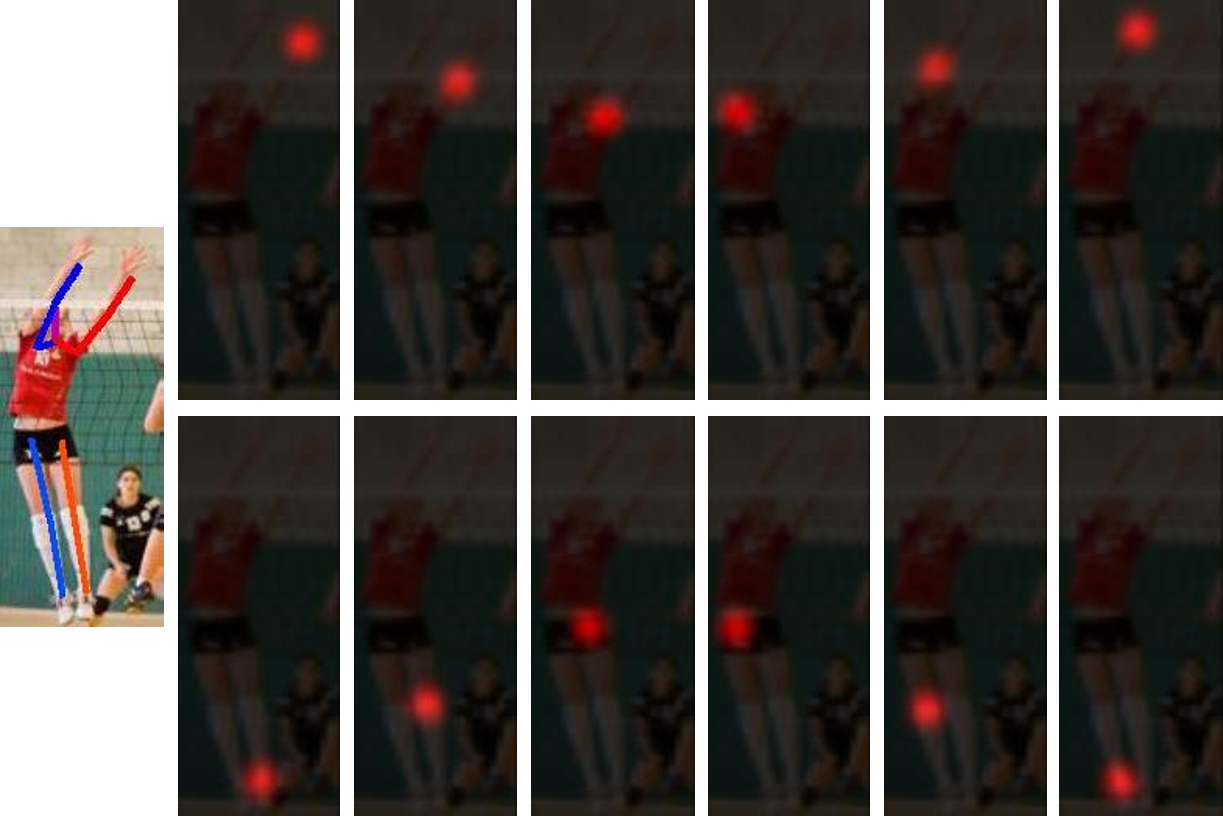} }
    \caption{\label{fig:hm}  Heatmaps visualization on the LSP dataset. (a) The predictions produced by the stacked hourglass network \cite{NewellYD16} are mostly accurate, but the heatmaps show that the model is not very sure about its answers. (b) Our method further refines the heatmaps and corrects the position of right shoulder. The heatmaps from left to right, top to bottom are left wrist, left elbow, left shoulder, right shoulder, right elbow, right wrist, left ankle, left knee, left hip, right hip, right knee, and right ankle.}
    \label{fig:improved_heatmaps}
 \end{figure*}

We show in Fig.~\ref{fig:qualitative} some qualitative results obtained using our method. Fig.~\ref{fig:improved_heatmaps} shows a visualization of heatmaps. It can be seen in Fig.~\ref{fig:improved_heatmaps_a} that the predictions produced by the stacked hourglass network \cite{NewellYD16} are mostly accurate, but the model is not very sure about its answers according to the heatmaps. Our method is able to refine the heatmaps, as shown in Fig.~\ref{fig:improved_heatmaps_b}.

\begin{itemize}
\item{LSP}:
The comparisons between our results and others are reported in Table~\ref{table:LSP}. Our model shown in this table is trained with external data from the MPII training set. The score is computed at $r = 0.2$. 
As shown in Fig.~\ref{fig:LSP-PCK}, our approach gets the highest detection rate across all tolerance range. Furthermore, the improvement is even more obvious at tighter distance (between $0.05$ and $0.1$).

\begin{table*}[tb]
    \centering
    \caption{Human pose estimation on the LSP dataset. (PCK)}
  		\begin{tabular}{|c||c|c|c|c|c|c|c||c|c|} \hline
		Methods &Head & Sho. & Elb. & Wri. & Hip & Knee  & Ank. & Total \\  \hline \hline
         Lifshitz \etal \cite{LifshitzFU16}, ECCV'16& 97.8  & 93.3  & 85.7  & 80.4  & 85.3  & 76.6 & 70.2 & 85.0 \\
         Pishchulin \etal\cite{PishchulinITAAG16} , CVPR'16& 97.0  & 91.0  & 83.8  & 78.1  & 91.0  & 86.7 & 82.0 & 87.1 \\
		Insafutdinov \etal \cite{InsafutdinovPAA16}, ECCV'16& 96.8  & {\bf 95.2}  & 89.3  & 84.4  & 88.4  & 83.4 & 78.0 & 88.5  \\
		Wei \etal \cite{WeiRKS16}, CVPR'16& 97.8  & 95.0  & 88.7  & 84.0  & 88.4  & 82.8 & 79.4 & 88.5  \\
        Bulat \etal\cite{BulatT16}, ECCV'16& 97.2  & 92.1  & 88.1  & 85.2  & 92.2  & 91.4 & 88.7 & 90.7 \\
		Chu \etal\cite{ChuYOMYW17}, CVPR'17& 98.1  & 93.7  & 89.3  & 86.9  & 93.4  & 94.0 & 92.5 & 92.6  \\ \hline \hline
    	Ours & {\bf 98.2}  & 94.9  & {\bf 92.2}  & {\bf 89.5}  & {\bf 94.2}  & {\bf 95.0} & {\bf 94.1} & {\bf 94.0} \\ \hline  
    	\end{tabular}    
    \label{table:LSP}
\end{table*}

\begin{figure*}[t]
\begin{center}
\begin{tabular}{@{}c@{}c@{}c@{}}
\includegraphics[width=0.35\textwidth]{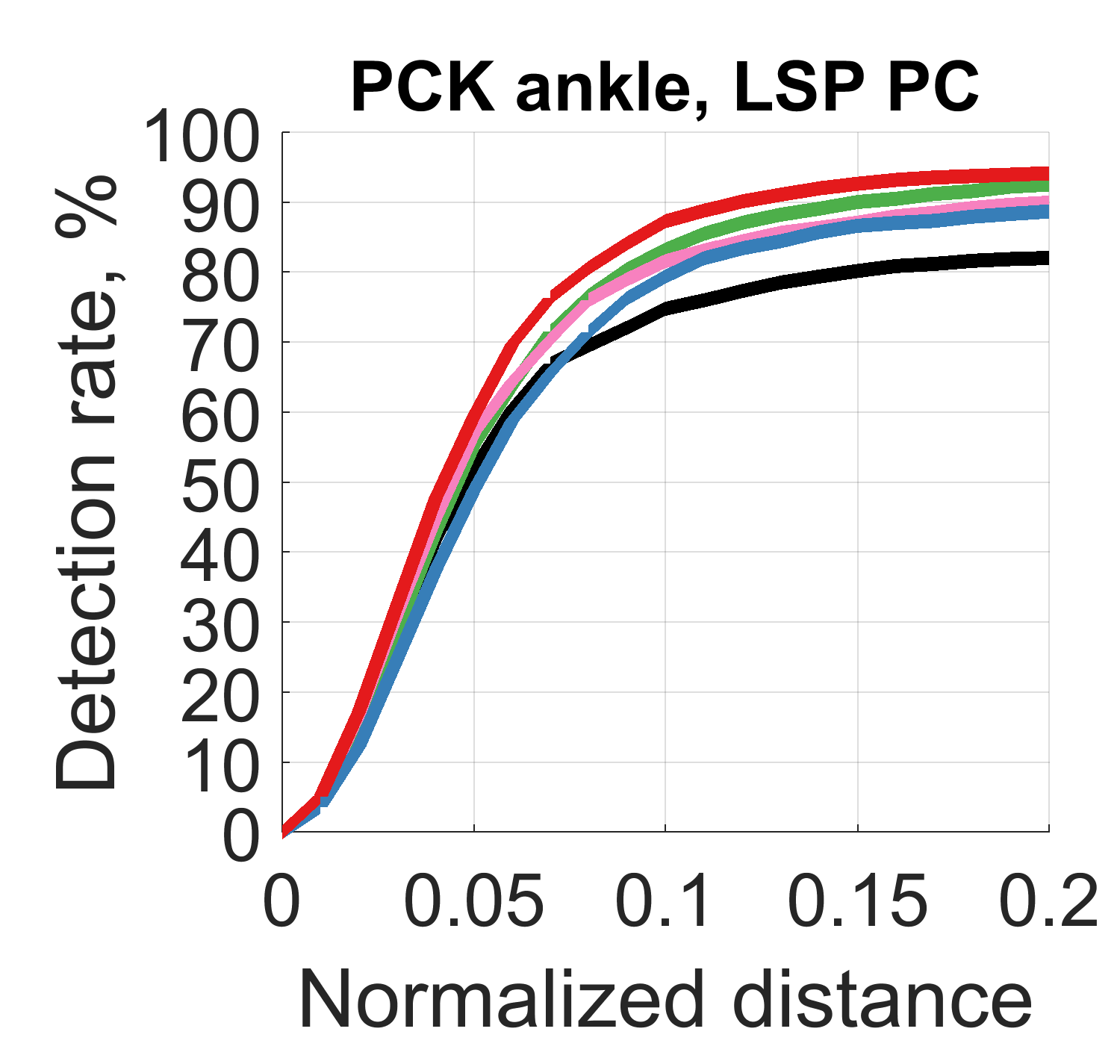} &
\includegraphics[width=0.35\textwidth]{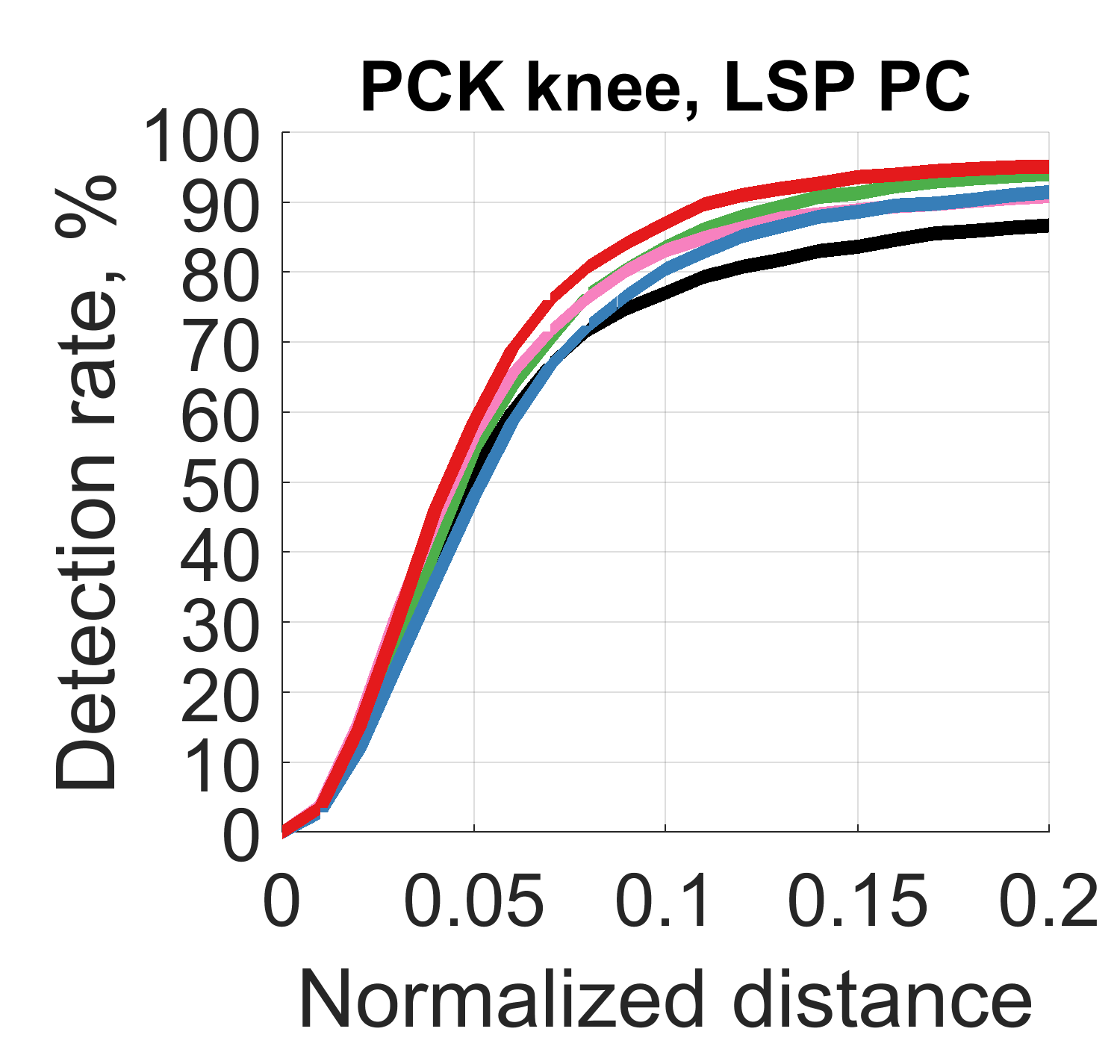} &
\includegraphics[width=0.35\textwidth]{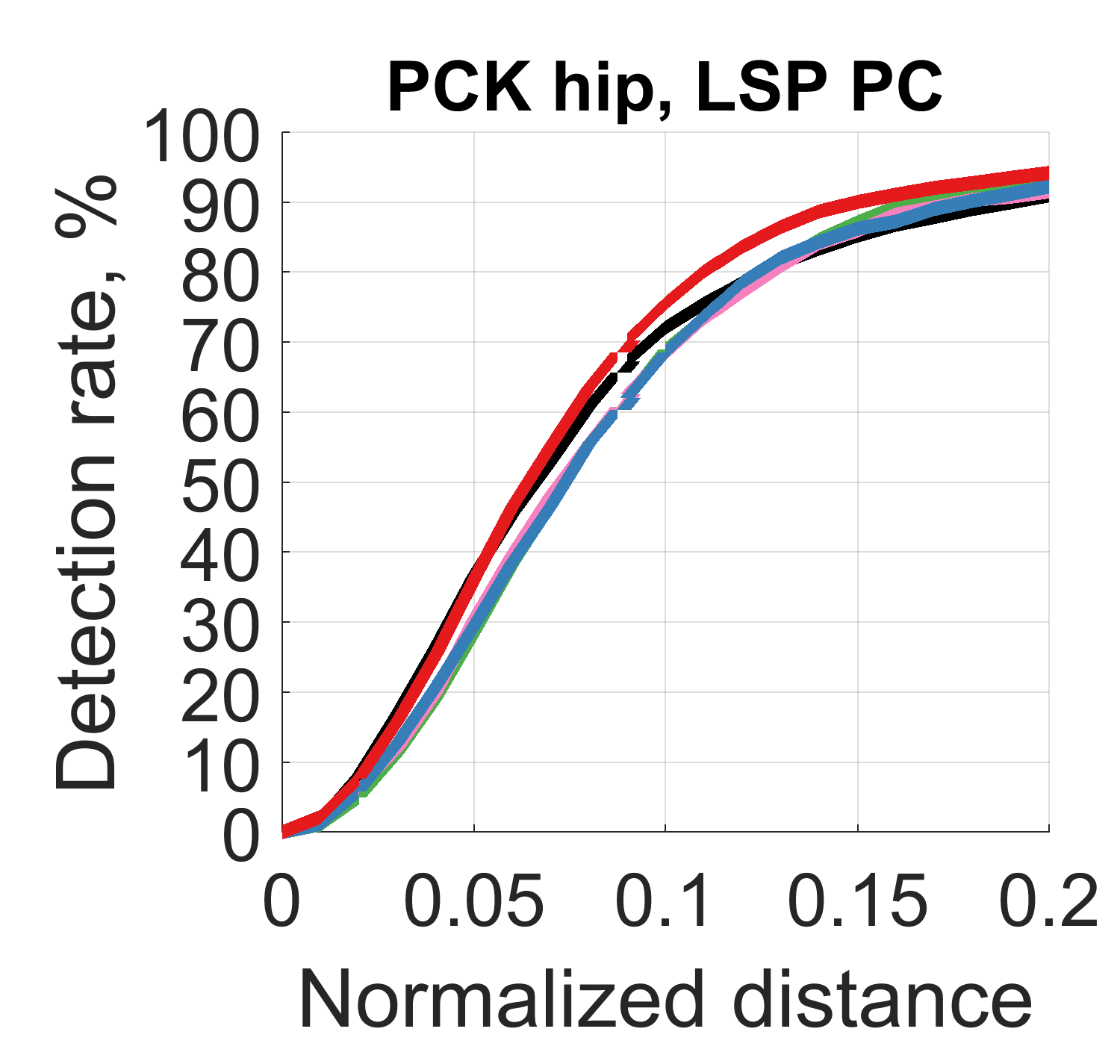} 
\\
\includegraphics[width=0.35\textwidth]{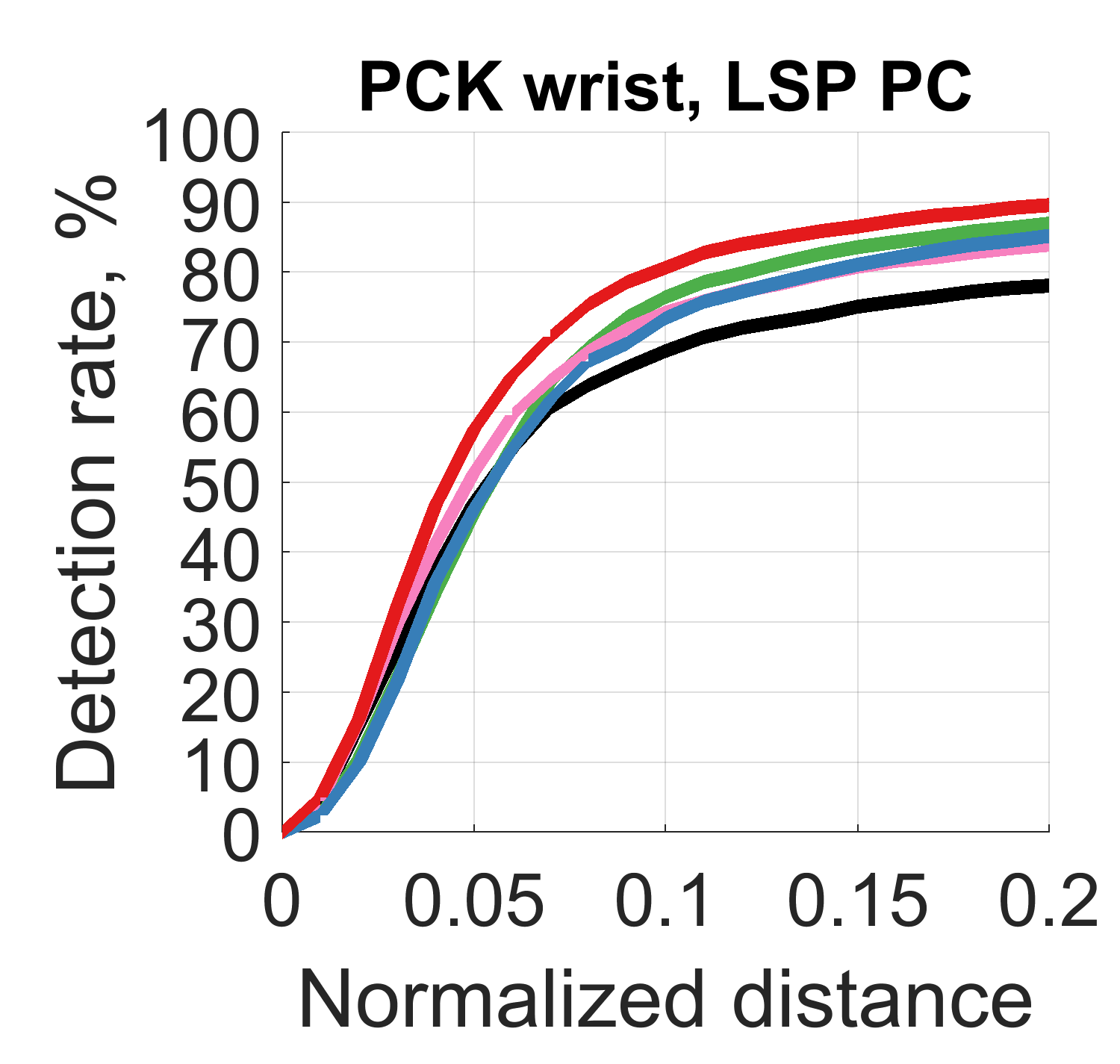} &
\includegraphics[width=0.35\textwidth]{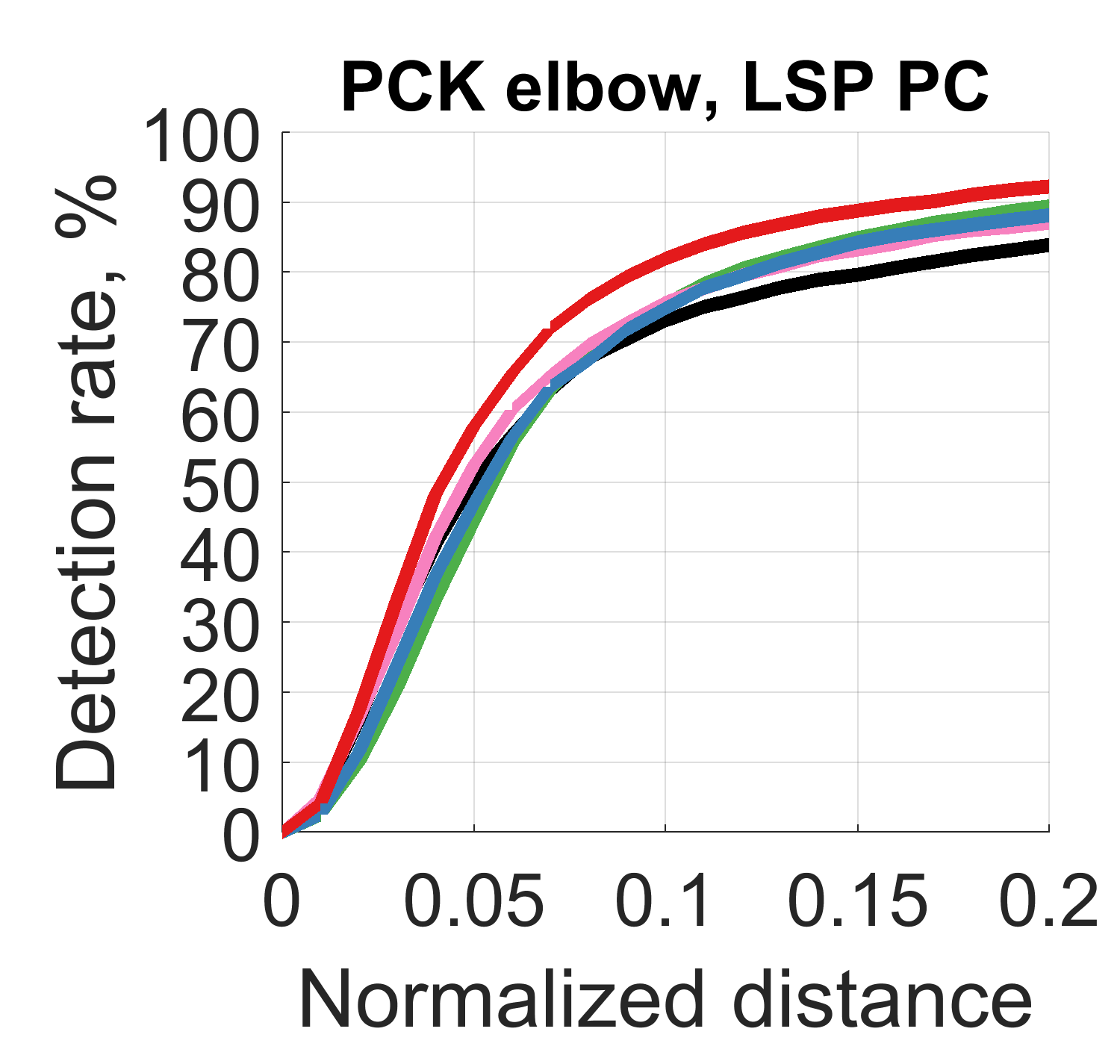} & 
\includegraphics[width=0.35\textwidth]{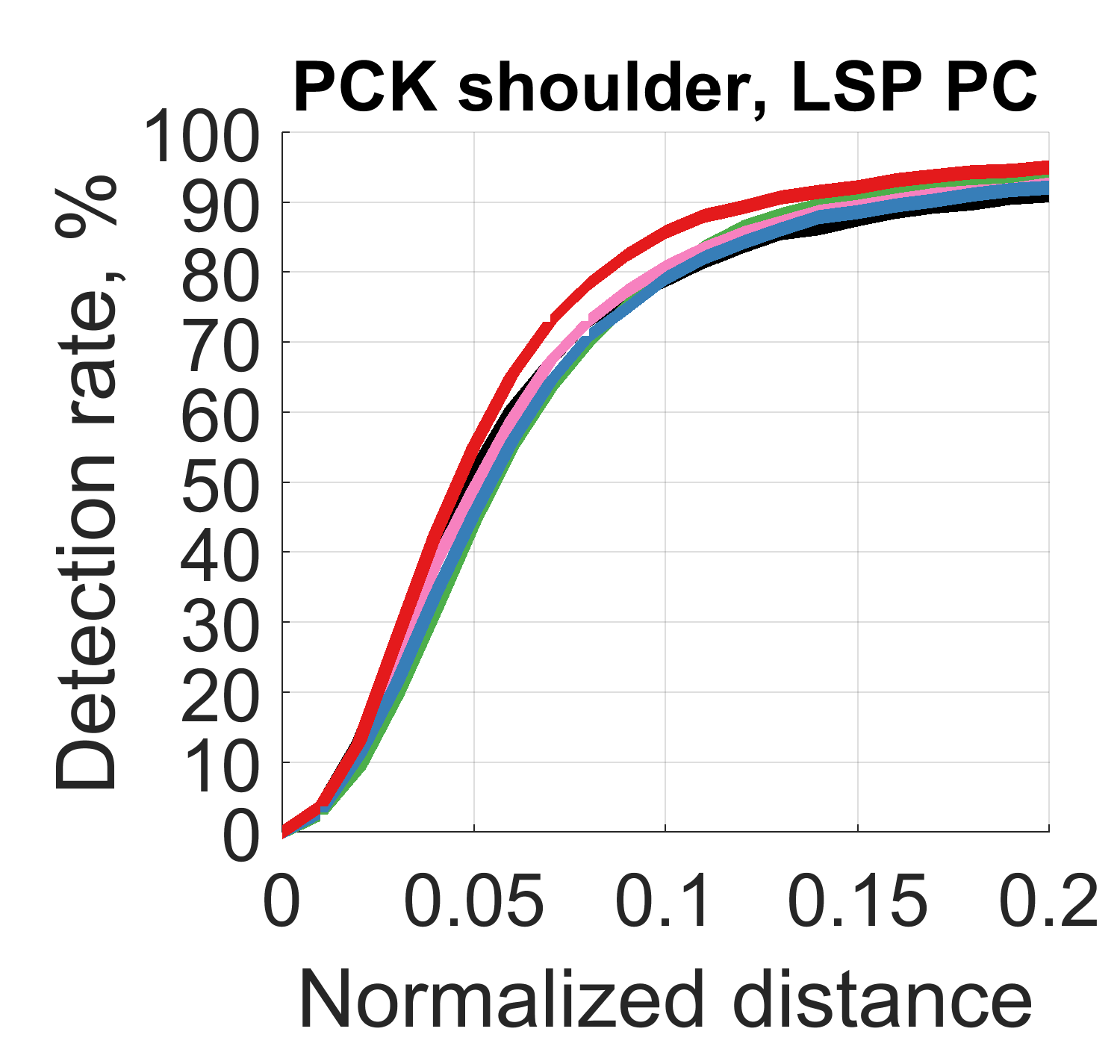} 
\\
\includegraphics[width=0.35\textwidth]{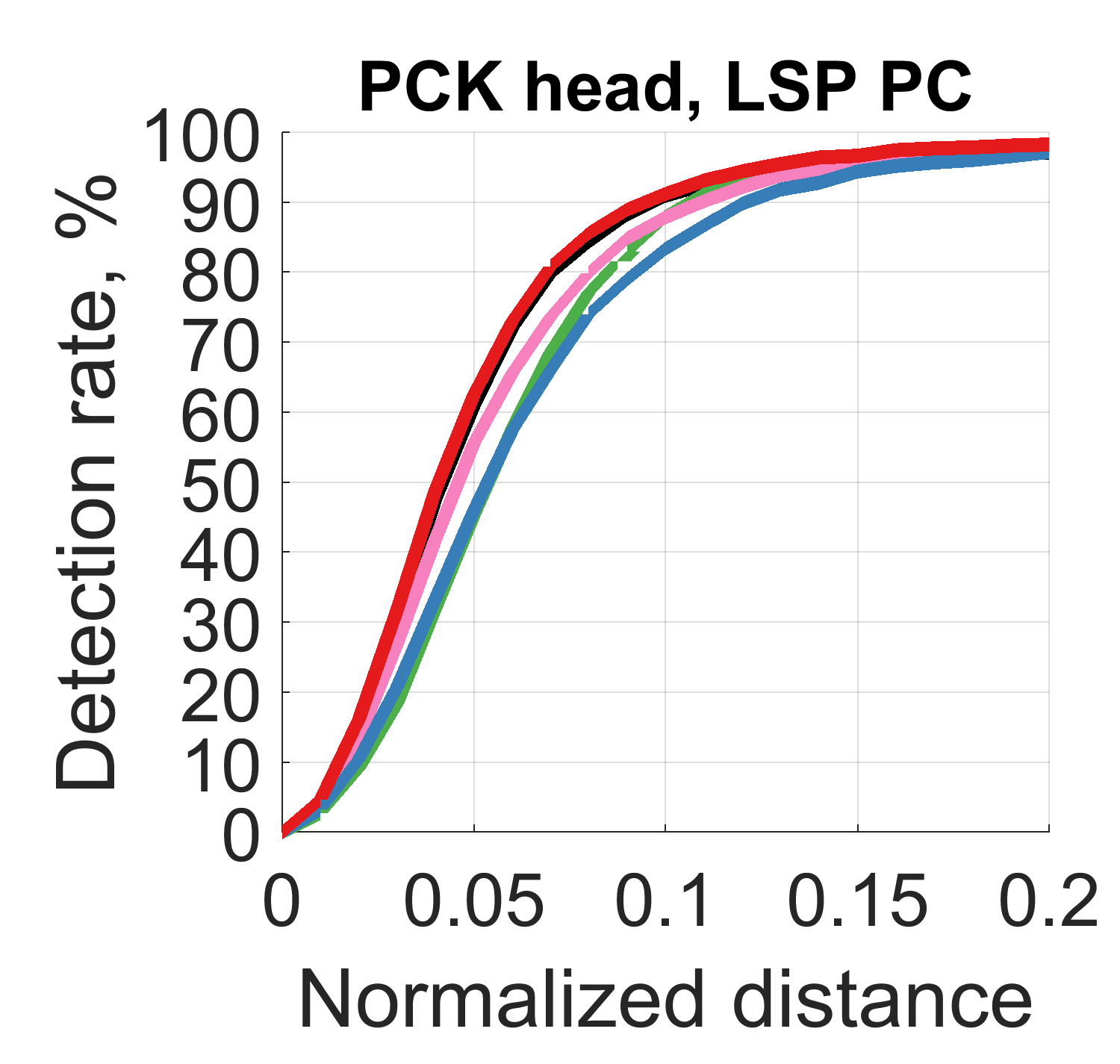} & 
\includegraphics[width=0.35\textwidth]{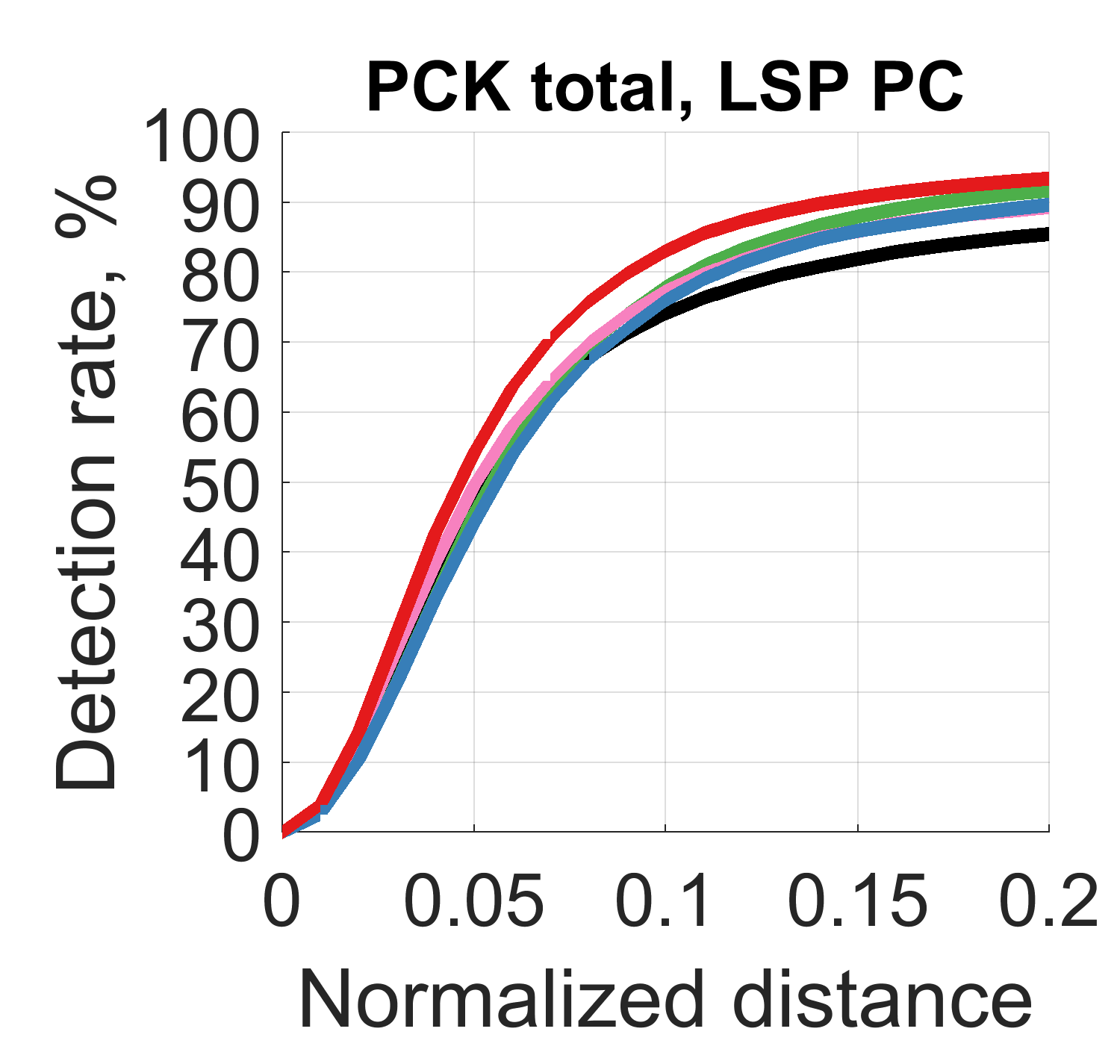} &
\includegraphics[width=0.3\textwidth]{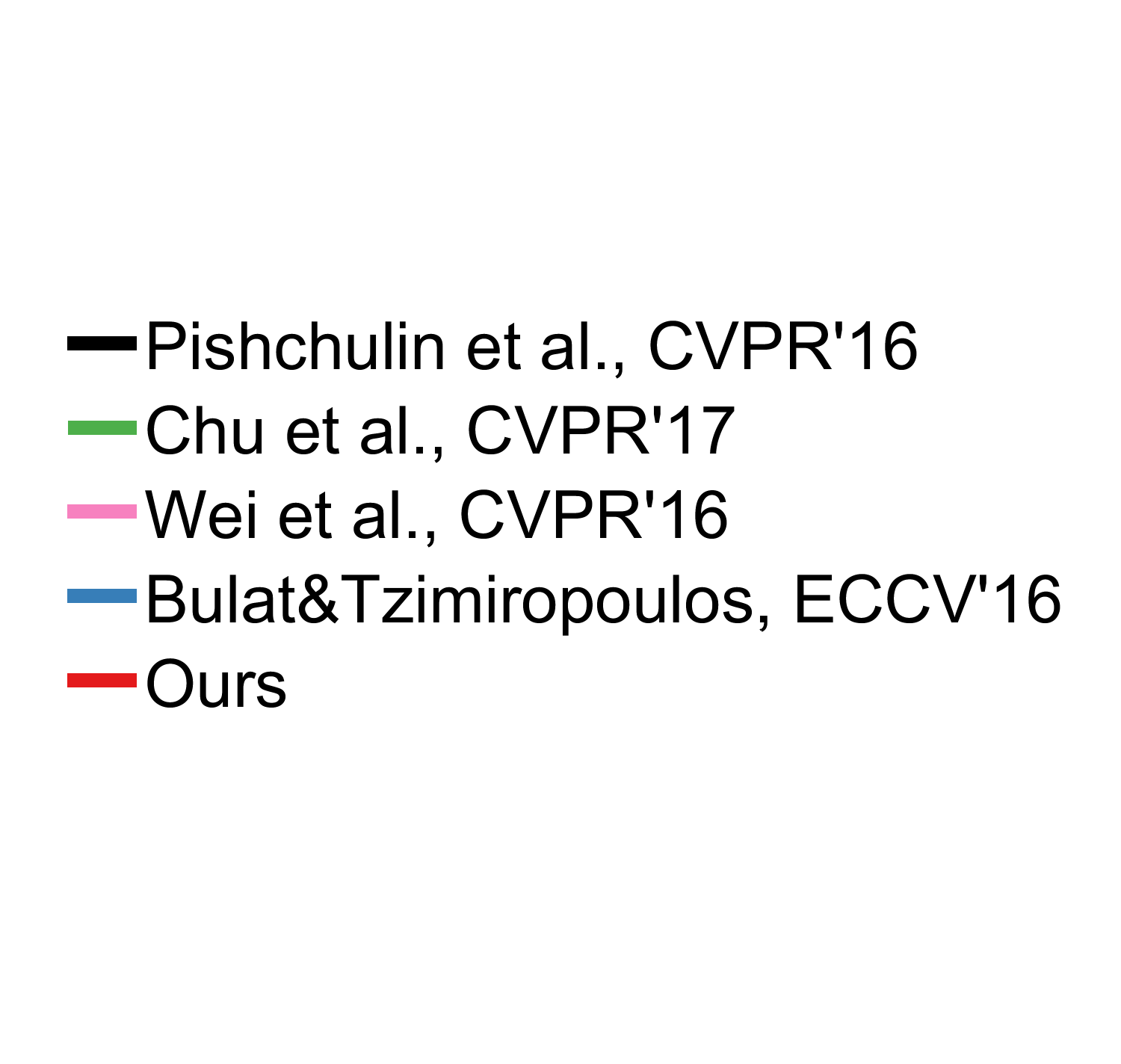}
\end{tabular}
   \caption{Percentage of Correct Keypoints (PCK) on the LSP dataset. All methods are trained with external data from the MPII training set, in addition to the LSP training set. PC refers to the person-centric annotation.}
\end{center}
\label{fig:LSP-PCK}
\end{figure*}

\item{MPII}:
Table~\ref{table:MPII} shows the PCKh performance of our method and previous methods at $r = 0.5$. Our model shown in this table is trained with external data from the LSP training set.

\begin{table*}[tb]
    \centering
    \caption{Human pose estimation on the MPII dataset. (PCKh)}
  		\begin{tabular}{|c||c|c|c|c|c|c|c||c|c|} \hline
		Methods &Head & Sho. & Elb. & Wri. & Hip & Knee  & Ank. & Total \\  \hline \hline
        Pishchulin \etal \cite{PishchulinAGS13}, ICCV'13& 74.3  & 49.0  & 40.8  & 34.1  & 36.5  & 34.4 & 35.2 & 44.1 \\ 
        Tompson \etal\cite{TompsonJLB14}, NIPS'14& 95.8  & 90.3  & 80.5  & 74.3  & 77.6  & 69.7 & 62.8 & 79.6  \\
        Carreira \etal\cite{CarreiraAFM15}, CVPR'16& 95.7  & 91.7  & 81.7  & 72.4  & 82.8  & 73.2 & 66.4 & 81.3  \\
        Tompson \etal\cite{TompsonGJLB15}, CVPR'15& 96.1  & 91.9  & 83.9  & 77.8  & 80.9  & 72.3 & 64.8 & 82.0  \\
        Hu \etal\cite{HuR16}, CVPR'16& 95.0  & 91.6  & 83.0  & 76.6  & 81.9  & 74.5 & 69.5 & 82.4  \\
        Pishchulin \etal\cite{PishchulinITAAG16}, CVPR'16& 94.1  & 90.2  & 83.4  & 77.3  & 82.6  & 75.7 & 68.6 & 82.4 \\
        Lifshitz \etal\cite{LifshitzFU16}, ECCV'16& 97.8  & 93.3  & 85.7  & 80.4  & 85.3  & 76.6 & 70.2 & 85.0  \\
        Gkioxary \etal\cite{GkioxariTJ16}, ECCV'16& 96.2  & 93.1  & 86.7  & 82.1  & 85.2  & 81.4 & 74.1 & 86.1  \\
        Rafi \etal\cite{RafiLGK16}, BMVC'16& 97.2  & 93.9  & 86.4  & 81.3  & 86.8  & 80.6 & 73.4 & 86.3 \\
        Belagiannis \etal\cite{BelagiannisZ16}, FG'17& 97.7  & 95.0  & 88.2  & 83.0  & 87.9  & 82.6 & 78.4 & 88.1  \\
        Insafutdinov \etal\cite{InsafutdinovPAA16}, ECCV'16& 96.8  & 95.2  & 89.3  & 84.4  & 88.4  & 83.4 & 78.0 & 88.5  \\
        Wei \etal\cite{WeiRKS16}, CVPR'16& 97.8  & 95.0  & 88.7  & 84.0  & 88.4  & 82.8 & 79.4 & 88.5  \\
        Bulat \etal\cite{BulatT16}, ECCV'16& 97.9  & 95.1  & 89.9  & 85.3  & 89.4  & 85.7 & 81.7 & 89.7  \\
        Newell \etal\cite{NewellYD16}, ECCV'16& 98.2  & 96.3  & 91.2  & 87.1  & 90.1  & 87.4 & 83.6 & 90.9  \\
        Chu \etal\cite{ChuYOMYW17}, CVPR'17& {\bf 98.5}  & 96.3  & 91.9  & {\bf 88.1}  & 90.6  & 88.0 & {\bf 85.0} & 91.5  \\ \hline \hline
        Ours& 98.2  & {\bf 96.8}  & {\bf 92.2}  & 88.0  & {\bf 91.3}  & {\bf 89.1} & 84.9 & {\bf 91.8} \\ \hline  
    	\end{tabular}    
    \label{table:MPII}
\end{table*}

\begin{figure*}[t]
\begin{center}
\begin{tabular}{@{}c@{}c@{}c@{}}
\includegraphics[width=0.35\textwidth]{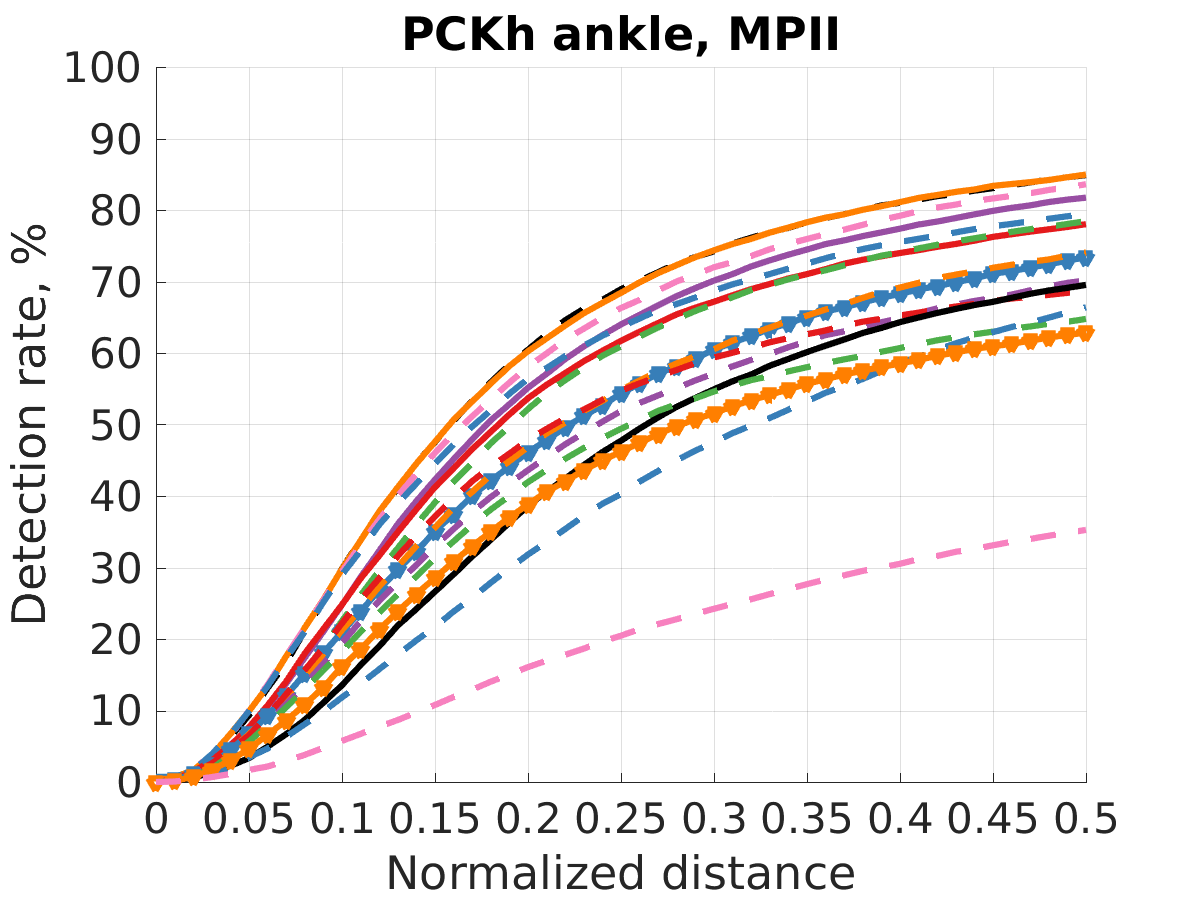} &
\includegraphics[width=0.35\textwidth]{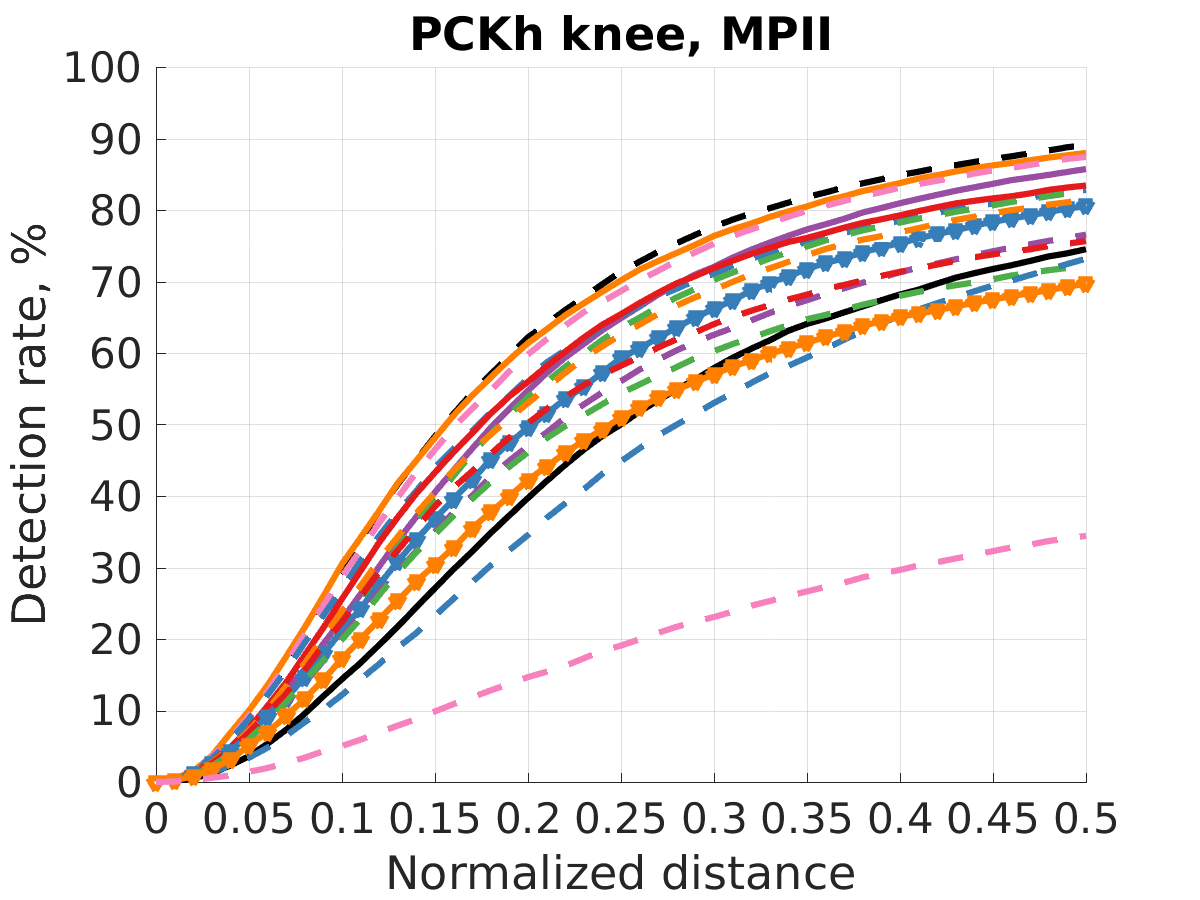} &
\includegraphics[width=0.35\textwidth]{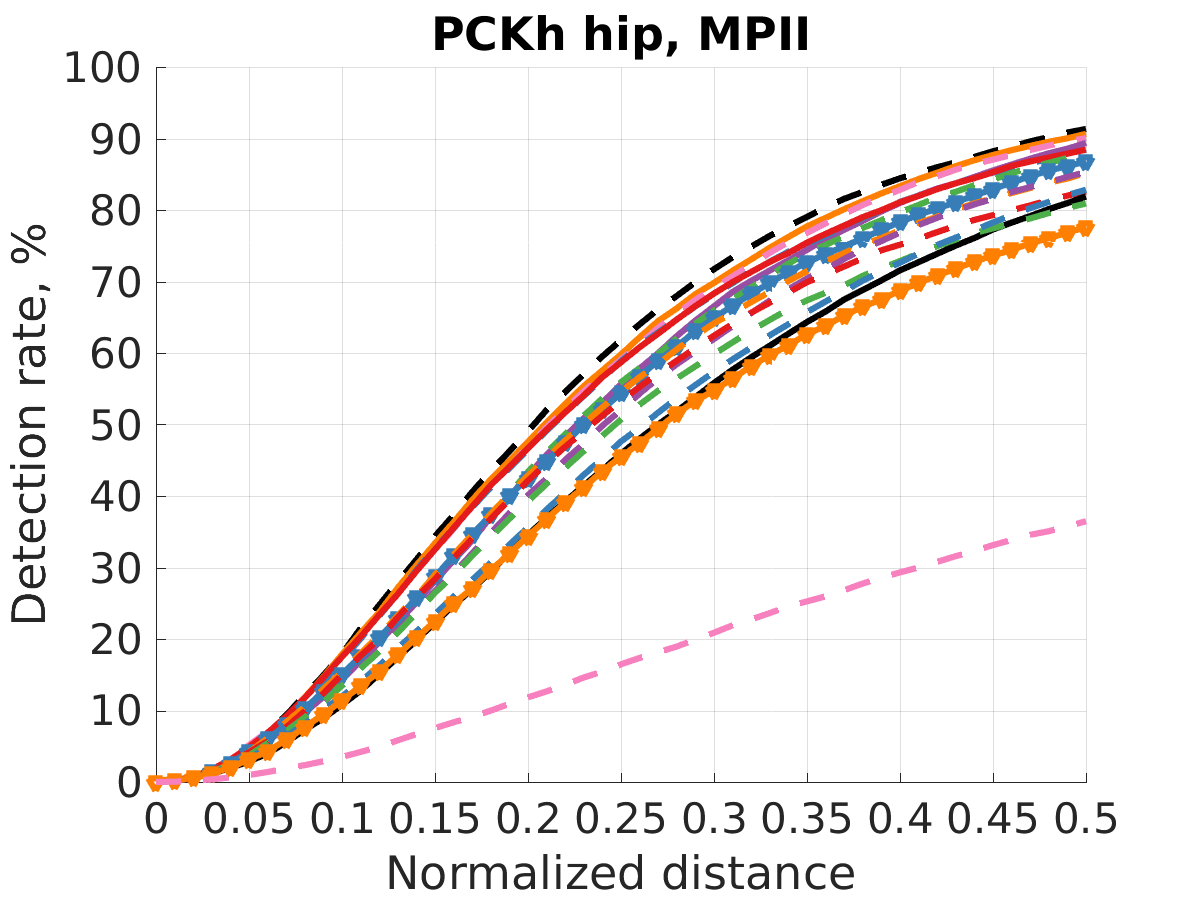} 
\\
\includegraphics[width=0.35\textwidth]{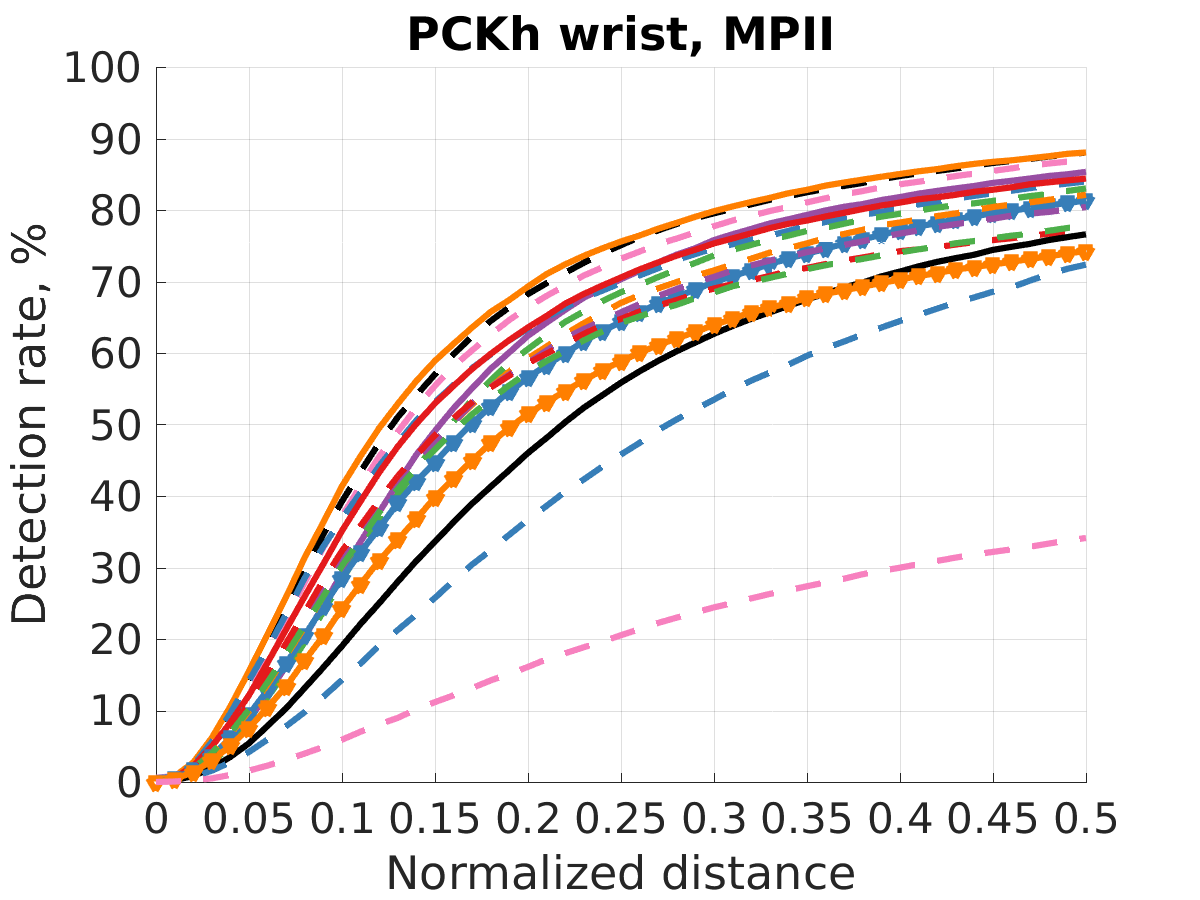} &
\includegraphics[width=0.35\textwidth]{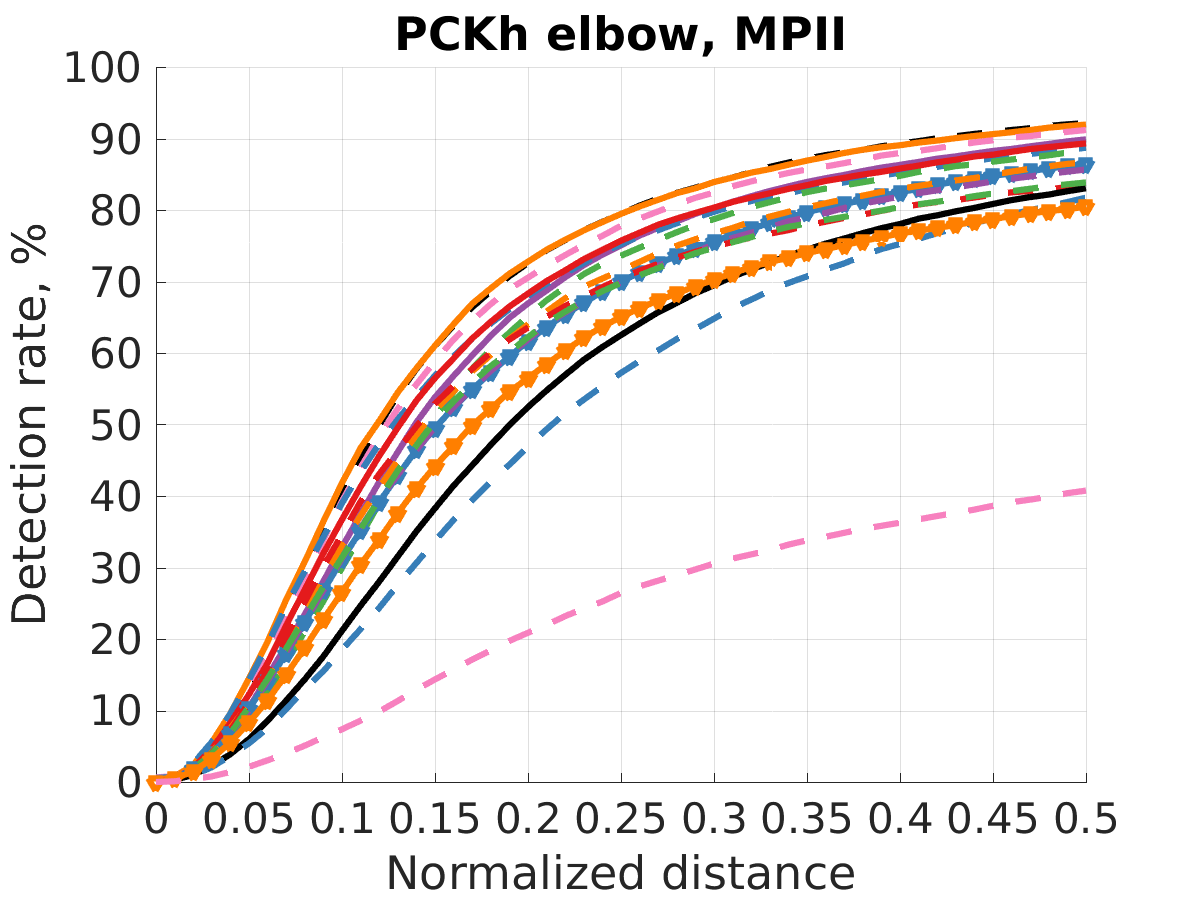} & 
\includegraphics[width=0.35\textwidth]{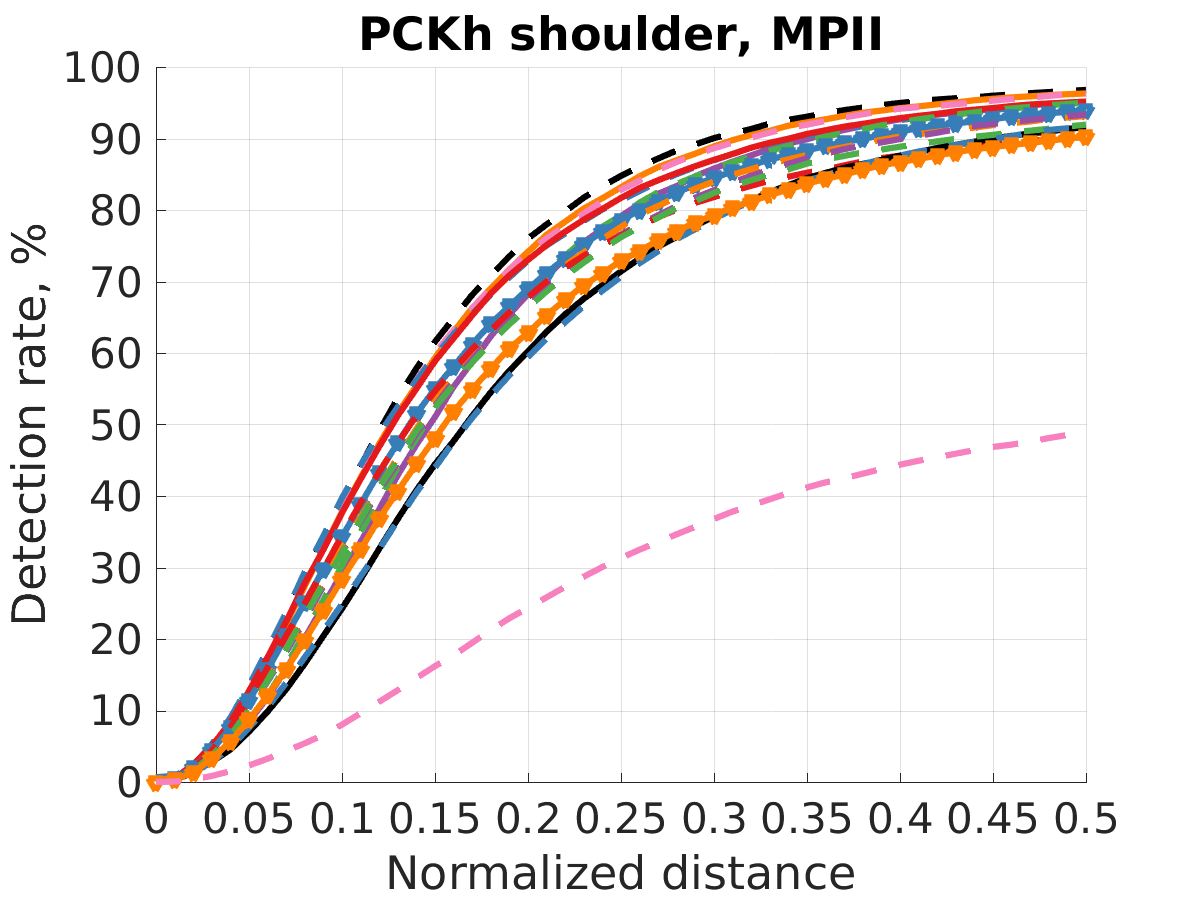} 
\\
\includegraphics[width=0.35\textwidth]{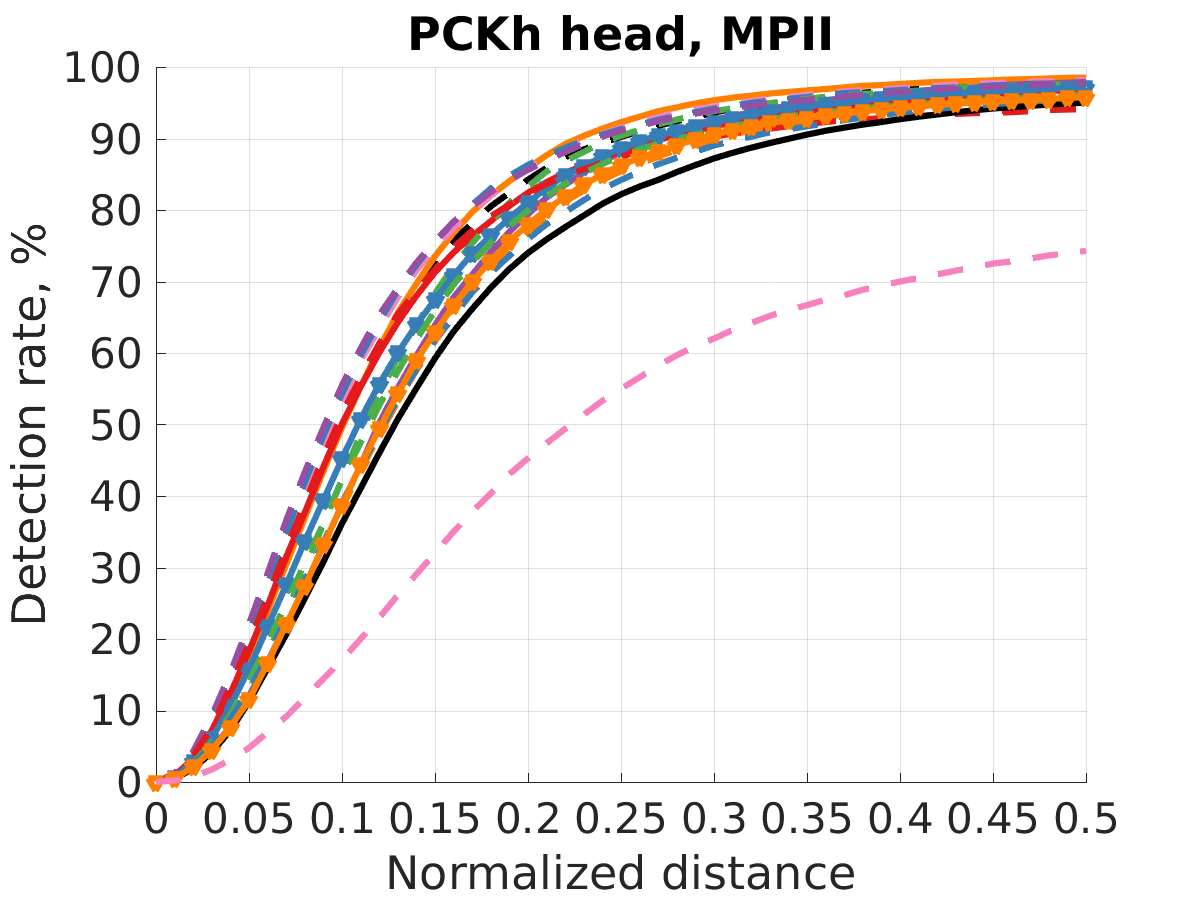} & 
\includegraphics[width=0.35\textwidth]{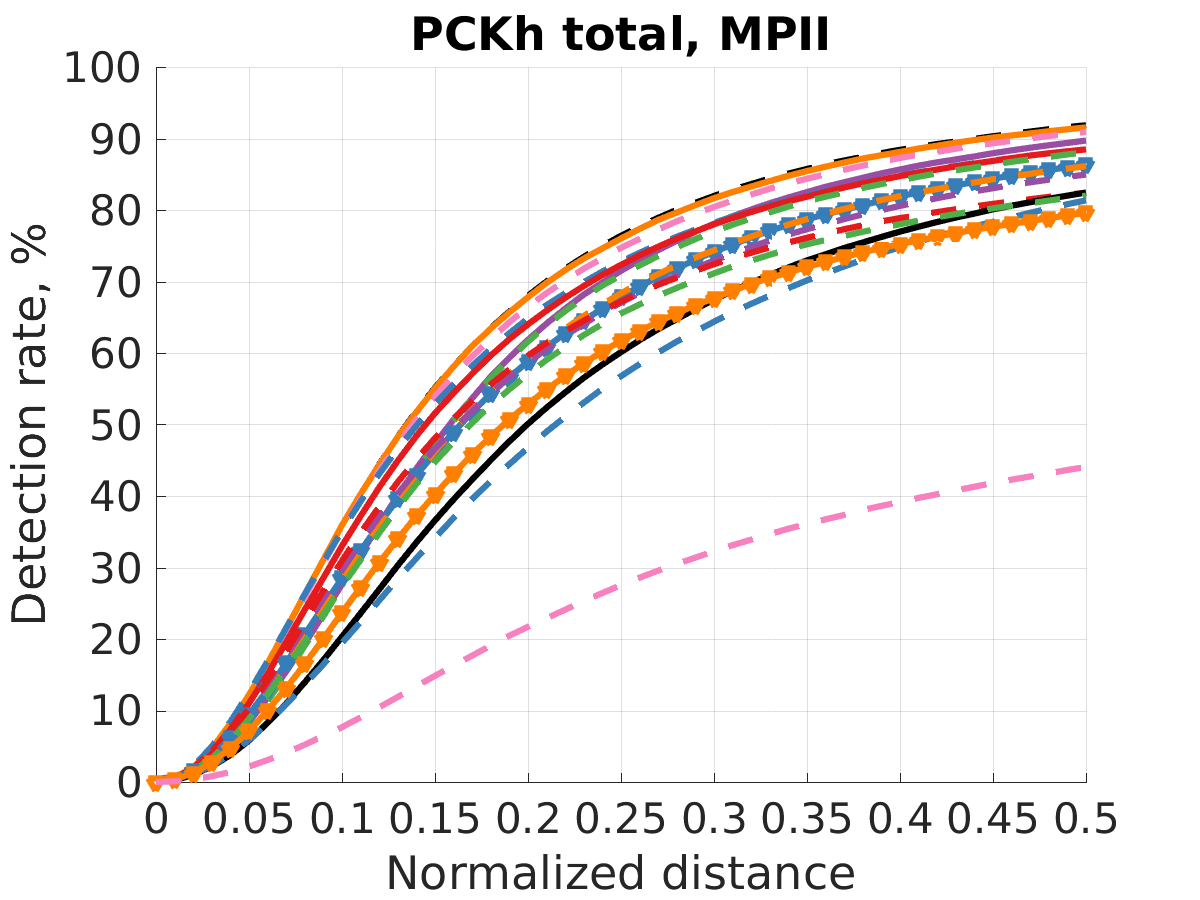} &
\includegraphics[width=0.35\textwidth]{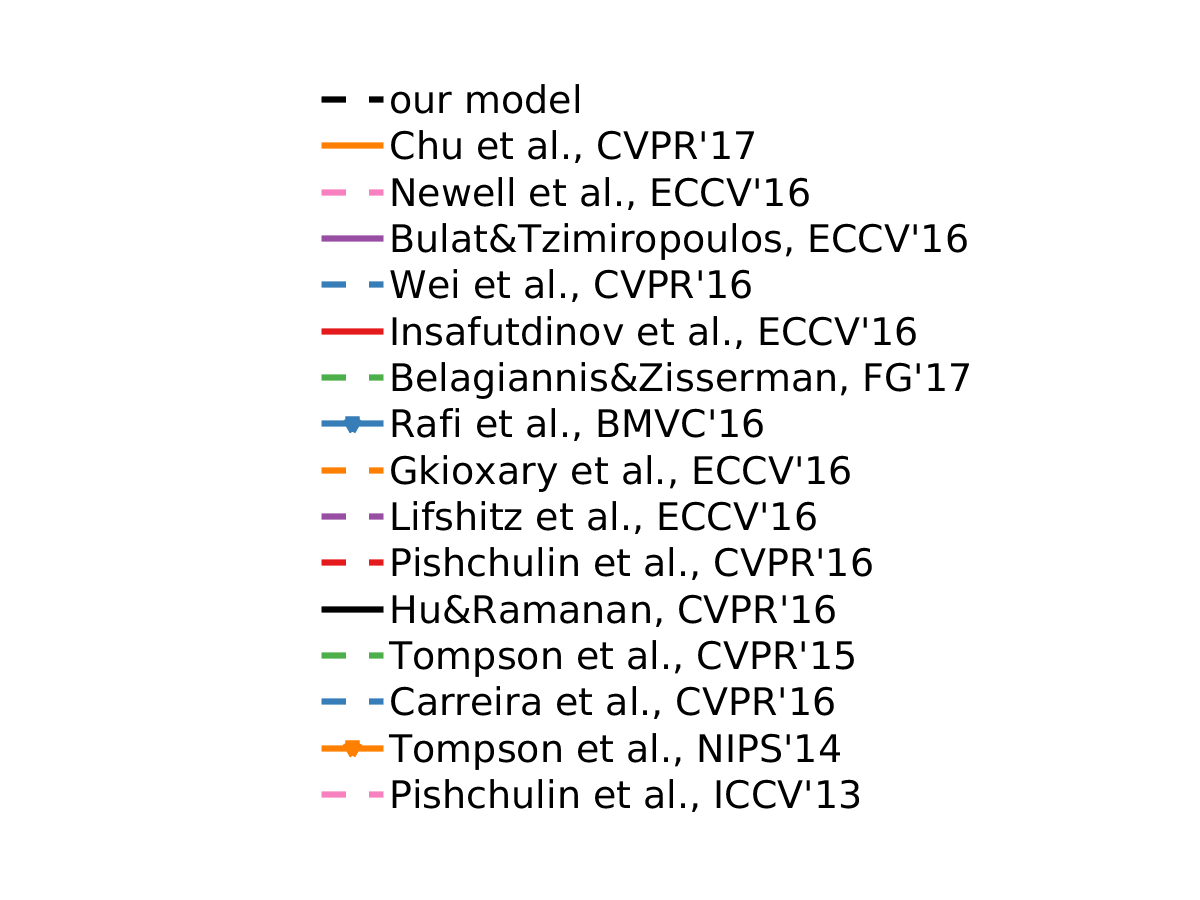}
\end{tabular}
   \caption{PCKh on the MPII dataset.}
\end{center}
\label{fig:MPII-PCK}
\end{figure*}

\item{LIP}:
Table~\ref{table:LIP} shows the final list of the CVPR 2017 LIP Human Pose Estimation Challenge. The challenge is finished and our method achieves the best result. For reference, both BUPTMM-POSE and Hybrid Pose Machine use methods that merge the predictions of Newell \etal\cite{NewellYD16} and others.

\end{itemize}

\begin{table*}[tb]
    \centering
    \caption{Human pose estimation on the LIP dataset. (PCK)}
  		\begin{tabular}{|c||c|c|c|c|c|c|c||c|c|} \hline
		Methods &Head & Sho. & Elb. & Wri. & Hip & Knee  & Ank. & Total \\  \hline \hline
        Hybrid Pose Machine& 71.7  & 87.1  & 82.3  & 78.2  & 69.2  & 77.0 & 73.5 & 77.2  \\
        BUPTMM-POSE& 90.4  & 87.3  & 81.9  & 78.8  & 68.5  & 75.3 & 75.8 & 80.2  \\
        Pyramid Stream Network& 91.1  & 88.4  & 82.2  & 79.4  & 70.1  & 80.8 & 81.2 & 82.1  \\ \hline \hline
        Ours& {\bf 94.9}  & {\bf 93.1}  & {\bf 89.1}  & {\bf 86.5}  & {\bf 75.7}  & {\bf 85.5} & {\bf 85.7} & {\bf 87.4} \\ \hline  
    	\end{tabular}    
    \label{table:LIP}
\end{table*}

\subsection{Analysis}

In this section, we present the effects of several components in our model. We conduct the experiments on the test set of the LSP dataset. We observe the accuracy through training iterations. 

\subsubsection{GAN and Conditional GAN}

\begin{figure*}[t]
\centering
    \subfigure[] { 
    \label{fig:analy1_a}
    \includegraphics[height=0.35\textwidth]{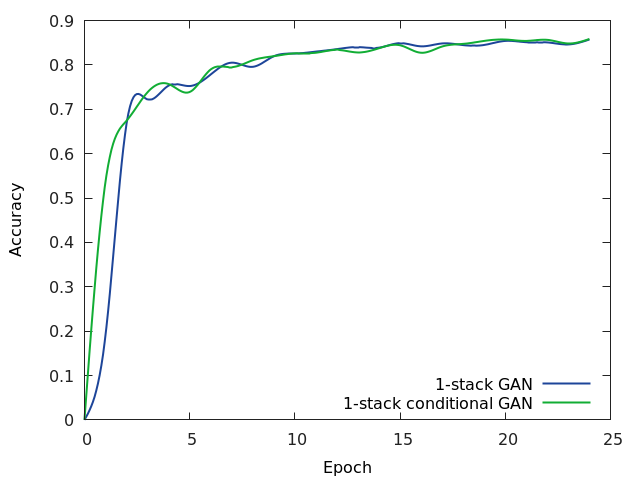} }
    \subfigure[] { 
    \label{fig:analy1_b}
    \includegraphics[height=0.35\textwidth]{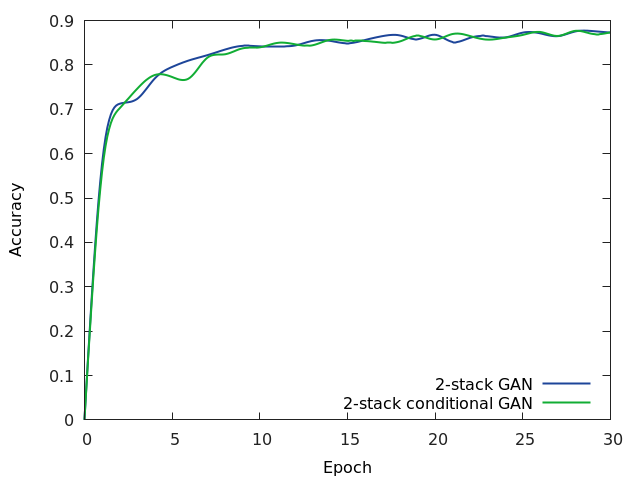} }
    \caption{\label{fig:analy1} PCK on the LSP dataset. The blue line is the accuracy of GAN while the green line is of conditional GAN. (a) 1-stack hourglass. (b) 2-stack hourglass. }
 \end{figure*}

We experiment on several network configurations. The settings differ in the number of stacks of the generator. The size of the discriminator is fixed (1-stack). As shown in Fig.~\ref{fig:analy1}, we find that GAN and conditional GAN perform almost equally in both 1-stack (Fig.~\ref{fig:analy1_a}) and 2-stack (Fig.~\ref{fig:analy1_b}). The discriminator seems to perform well even when the image of the person is not provided. A possible reason is that the implausible pose could be recognized by merely the pose information. The image of the person is an extra information, but the discriminator does not always need it.

\subsubsection{With or without Adversarial Training}

\begin{figure*}[t]
\centering
    \subfigure[] {
    \label{fig:analy2_a}
    \includegraphics[height=0.35\textwidth]{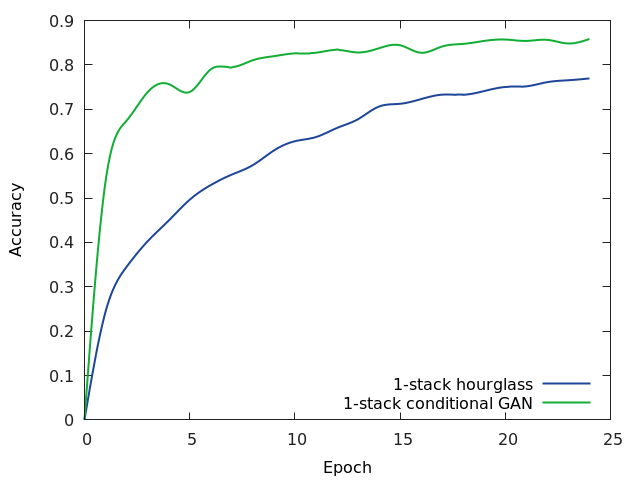} }
    \subfigure[] { 
    \label{fig:analy2_b}
    \includegraphics[height=0.35\textwidth]{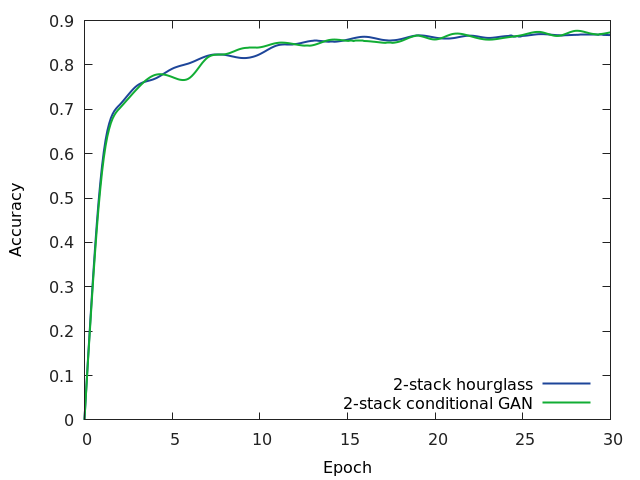} }
    \subfigure[] { 
    \label{fig:analy2_c}
    \includegraphics[height=0.35\textwidth]{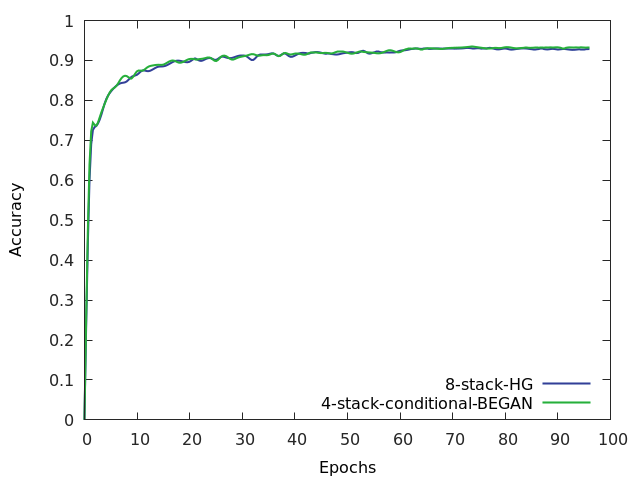} }
    \subfigure[] { 
    \label{fig:analy2_d}
    \includegraphics[height=0.35\textwidth]{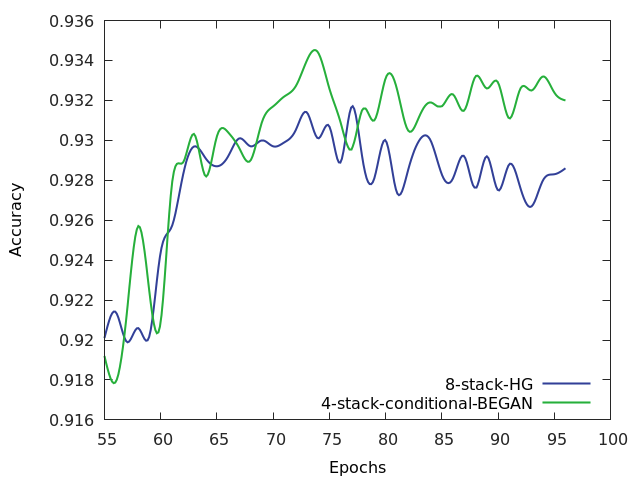} }
    \caption{\label{fig:analy2} PCK on the LSP dataset. The blue line is the approach of \cite{NewellYD16} while the green line is ours.  (a) 1-stack hourglass. (b) 2-stack hourglass.
    (c) 8-stack standard hourglass versus 4-stack hourglass plus a discriminator. In this setting we decrease the learning rate by $10^{-1}$ at epoch 60. (d) We zoom in the part of curve after epoch 60. We find that the strategy of learning rate decay is helpful for both methods, but ours is a bit more stable and achieves better performance in the end.}
 \end{figure*}

To investigate the benefit of adversarial training, we compare our method with the original stacked hourglass network. In Fig.~\ref{fig:analy2_a}, the improvement of adding adversarial training is significant. Our method converges faster and ends at a higher accuracy. But when it comes to 2-stack hourglass, in Fig.~\ref{fig:analy2_b}, the gain of adversarial training does not seem so obvious like 1-stack hourglass. The lines are staggered across training iterations, although at the end our method is a little higher than the original hourglass. The 8-stack hourglass is the best setting released by the authors of \cite{NewellYD16}, but in our experiment, in Fig.~\ref{fig:analy2_c}, 4-stack hourglass plus a discriminator is a better choice. In this setting we decrease the learning rate by $10^{-1}$ at epoch 60. In Fig.~\ref{fig:analy2_d}, we zoom in the part of curve after epoch 60. We find that the strategy of learning rate decay is helpful for both methods, but ours is a bit more stable and achieves better performance in the end.

\section{Conclusion}

We present an adversarial network to solve the human pose estimation problem. The network is composed of a generator and a discriminator with the same architecture. The generator is responsible for predicting the heatamps of human body keypoints based on the image features, and the discriminator plays the role of critic that can distinguish implausible poses and give the generator useful hints to improve the heatmaps. The additional discriminator can be removed after the training is done, and therefore it does not affect the inference time. We evaluate our approach on three standard benchmark datasets and the results show that our approach is useful for improving the prediction accuracy.

\bibliography{vision}
\bibliographystyle{ieee}

\end{document}